\documentclass[%
 reprint,
 amsmath,amssymb,
 aps,
]{revtex4-2}

\usepackage{graphicx}% Include figure files
\usepackage{dcolumn}% Align table columns on decimal point
\usepackage{bm}% bold math
\usepackage{array}
\usepackage{algpseudocode}
\usepackage{xcolor}
\newcommand{\changed}[1]{\textcolor{black}{#1}}
\begin{document}

\preprint{APS/123-QED}

\title{Towards agent-based-model informed neural networks}% Force line breaks with \\

\author{Nino Antulov-Fantulin}
\email{nino@aisot.ch, anino@ethz.ch}
\affiliation{%
Aisot Technologies AG, R\&D\\
 ETH Zurich, Computational Social Science
}%

\date{\today}% It is always \today, today,

\maketitle

\section*{Abstract}
In this article, we present a framework for designing neural networks that remain consistent with the underlying 
principles of agent-based models. 
\changed{We begin by highlighting the limitations of standard neural differential equations in modeling complex systems, 
where physical invariants (like energy) are often absent but other constraints (like mass conservation, information locality, bounded rationality) must be enforced.
To address this, we introduce \emph{Agent-Based-Model informed Neural Networks} (ABM-NNs), which leverage restricted graph neural networks and 
hierarchical decomposition to learn interpretable, structure-preserving dynamics.
We validate the framework across three case studies of increasing complexity:
(i) a Generalized Lotka--Volterra system, where we recover ground-truth parameters from short trajectories in presence of interventions;
(ii) a graph-based SIR contagion model, where our method outperforms state-of-the-art graph learning baselines (GCN, GraphSAGE, Graph Transformer) in out-of-sample forecasting and noise robustness; and
(iii) a real-world macroeconomic model of the ten largest economies, where we learn coupled GDP dynamics from empirical data and demonstrate counterfactual analysis for policy interventions.}

\textbf{Keywords:} Agent-based models, Neural ordinary differential equations, 
Graph neural networks, Complex systems, Network dynamics

\section*{Introduction}
\begin{figure}[htbp]
    \centering
    \includegraphics[width=0.98\linewidth]{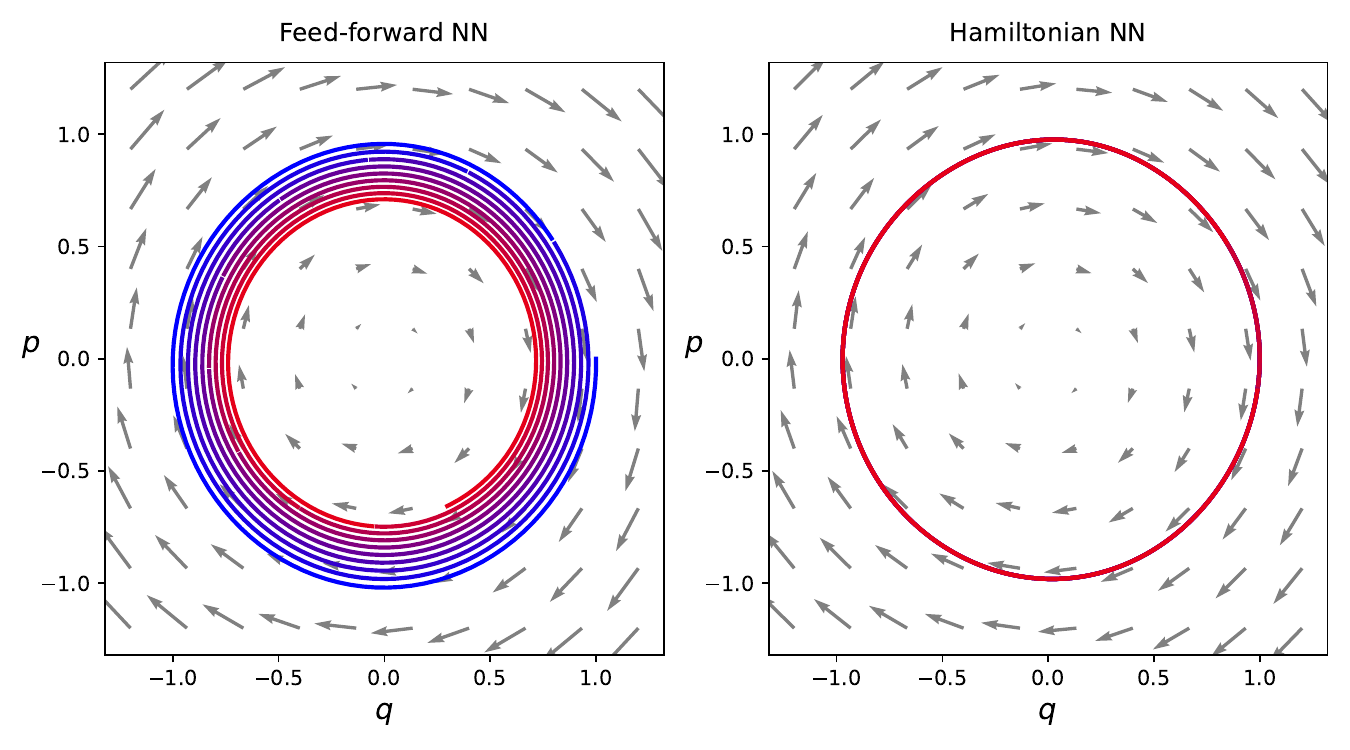}
    % Generated by: python hnn_demo.py (kept as a note, file name omitted from caption)
    \caption{Comparison between a feed-forward neural network (left) and a physics-informed Hamiltonian neural network (right)~\cite{Greydanus2019Hamiltonian} on an ideal spring system. 
    The HNN preserves energy and yields structure-consistent trajectories, while the unconstrained network can violate physical invariants. \changed{The color gradient from blue to red 
    denotes the progression of time from the initial time to the final integration time. }}
    \label{fig:neural-ode-vs-hnn}
\end{figure}

% story argument "introduction": Neural networks are universal approximators, but for modelling dynamical systems we need more tools in our box
 Neural networks (NN) are universal approximators~\cite{Hornik1989Universal,Cybenko1989Approximation}, but effective modeling of dynamical systems typically requires 
 inductive biases that encode structure (e.g., invariants, conservation laws, sparsity) for stability and generalization.
% story argument "introduction": RNNs & CTRNNs are universal approximators for modelling dynamical systems. add citations
  Recurrent neural networks and continuous-time RNNs are universal for representing dynamical behaviors~\cite{Schafer2007RNNUniversal,Funahashi1993CTRNN}, 
  yet unconstrained forms can violate physical constraints or overfit, motivating principled structure in the vector field.
% story argument "introduction": view1 on Neural ODEs: infinite depth Neural nets. add citation to neural ODEs
Recently, the Neural ODE framework was introduced~\cite{Chen2018NeuralODE}, where Neural ODEs can be viewed as the infinite-depth limit of residual networks,
 with continuous depth and parameters defining a flow~\cite{Chen2018NeuralODE}.
 \begin{equation}
    \frac{dh(t)}{dt} = \phi_\theta(h(t), t),
\end{equation}
where $\phi_\theta$ is a neural network with parameters $\theta$.
 % story argument "introduction": view2 on Neural ODEs: Neural ODEs approximate dynamics by saying: the world evolves according to some unknown vector field; let’s let a neural net represent it directly.
 Equivalently, they posit that the world evolves under an unknown vector field and directly parameterize this field with a neural network to be learned from data. 
% story argument "introduction": limitation of modelling with Neural ODE: mention augmented Neural ODEs and Hamiltonian Neural Networks. 
 Augmented neural ODEs expand the state to increase expressivity, enabling modeling of discontinuities, events, or inherently non-invertible mappings~\cite{Dupont2019AugmentedNODE}. \changed{Continuous-time sequence models such as neural controlled differential equations extend this family to irregularly sampled time series~\cite{Kidger2020NeuralCDE}.}
However, from a physics perspective, larger expressivity can lead to overfitting and violation of physical constraints.
% story argument "introduction": see figure \label{fig:neural-ode-vs-hnn}, which shows that Neural ODE can violate physical constraints.
 As illustrated in Fig.~\ref{fig:neural-ode-vs-hnn} left, unconstrained Neural ODEs may drift away from energy-conserving trajectories.
% story argument "hnn": Hamiltonian neural networks are a principled framework for learning dynamics from data while preserving the symplectic geometry of Hamiltonian mechanics.
Among structure-preserving approaches, \emph{Hamiltonian neural networks}~\cite{Greydanus2019Hamiltonian} (HNNs) provide a principled framework for learning dynamics from data while preserving the symplectic geometry of Hamiltonian mechanics (i.e., energy conservation). Instead of directly predicting trajectories, an HNN parameterizes the Hamiltonian function $H(q,p)$ in terms of generalized coordinates $q$ and conjugate momenta $p$. The time evolution then follows from Hamilton’s equations,
\begin{equation}
    \dot{q} = \frac{\partial H}{\partial p}, \qquad
    \dot{p} = -\frac{\partial H}{\partial q},
\end{equation}
ensuring energy conservation and structure-preserving predictions even over long time horizons. 
This inductive bias distinguishes HNNs from standard neural ODE frameworks, offering improved generalization and interpretability in modeling physical systems.
% story argument "hnn": see figure \label{fig:neural-ode-vs-hnn}, which shows that Hamiltonian neural networks can preserve energy conservation.
 See Fig.~\ref{fig:neural-ode-vs-hnn} right, which shows that Hamiltonian neural networks can preserve energy conservation,
 and thus physical realism. Analogously, one can encode physical reality via Lagrangian Neural Networks that parameterize the Lagrangian and derive dynamics from the Euler--Lagrange equations~\cite{Cranmer2020LNN}.
 \changed{Related physics-informed approaches constrain neural surrogates to satisfy governing equations, either through PDE residuals in physics-informed neural 
 networks~\cite{Raissi2019PINN} or through operator-learning architectures such as Fourier Neural Operators~\cite{Li2020FNO}, 
 and have been surveyed under the broader banner of physics-informed machine learning~\cite{Karniadakis2021PIML}.}
 %Alternative data-driven methods such as sparse identification of nonlinear dynamics (SINDy) recover explicit governing equations from trajectory data~\cite{Brunton2016SINDy}
 %by using pre-defined dictionary of functions.
% In the HNN panel, conservation of energy keeps the trajectory on a constant-energy orbit so successive colors overlap, whereas in the unconstrained feed-forward network the system loses energy and the colored trajectories drift away from the constant-energy curve.

% story argument " from classical mechanics to agent-based models" 
% Dynamical processes like in complex socio-techno-economic systems are usually composed out of large number of different
% interacting agents, which interact via complex network or graph and give rise to collective dynamics.
% Statistical mechanics models large numbers of particles by replacing the impossible task of tracking each one with a 
% probabilistic description over all possible microstates, and by assuming the system is in equilibrium, ensembles can be 
% used to average these probabilities and extract macroscopic properties like temperature, pressure, and entropy that emerge from collective behavior.
% however many real-world systems are not in equilibrium, and their dynamics is not governed by a single principle.
% that is why agent-based models are a natural fit for modeling complex systems. 

\changed{In order to model the dynamics of complex systems, we need to model the interactions of large number of heterogeneous agents, 
coupled through complex interaction patterns~\cite{Newman2010Networks}, and for this Lagrangian or Hamiltonian frameworks are not adequate.
Classical statistical mechanics~\cite{Sethna2006Statistical} models large numbers of particles by replacing the impossible task of 
tracking each one with a probabilistic description over all possible microstates, and by assuming the system is in equilibrium, 
ensembles can be used to average these probabilities and extract macroscopic properties like temperature, pressure,
and entropy that emerge from collective behavior.}

\changed{However, many socio-techno-economic systems operate far from equilibrium, and this is particularly evident in systems exhibiting 
self-organized criticality, where the endogenous drive 
for efficiency pushes the system to a fragile state at the edge of chaos, making large-scale fluctuations an intrinsic feature 
rather than an anomaly~\cite{bouchaud2024self}. 
To model the dynamics of socio-techno-economic systems that operate far from equilibrium, a diverse array of non-equilibrium frameworks has been developed, 
ranging from (i) stochastic frameworks, (ii) computational \& simulation frameworks, to (iii) macroscopic deterministic approximations.
\textit{Stochastic frameworks} provide the exact mathematical description of a system's evolution by tracking probability distributions, 
event timings, or network states without ignoring fluctuations. 
Central to this approach is the Master Equation, which describes the time-evolution of the probability of the system being in a specific discrete microstate 
and serves as the fundamental formalism for stochastic jump processes in continuous time~\cite{vankampen1992stochastic, gardiner2009stochastic}. 
Complementing this state-based view are Self-Exciting Point Processes, such as Hawkes Processes, which model the precise timing of events in continuous time. 
These capture how past events trigger ``avalanches'' of future activity, a feature crucial for modeling endogenous bursting in financial and 
social systems~\cite{hawkes1971spectra, bacry2015hawkes}. 
Additionally, for systems defined by sparse, heterogeneous networks, analytical techniques like Cavity Methods or Message Passing (e.g., Belief Propagation) provide tools 
to solve statistical mechanics problems by capturing how local topology influences marginal probabilities~\cite{mezard2009information, zdeborova2016statistical}.
Since exact solution to the Master Equation are often analytically insoluble for large complex systems, \textit{computational and simulation frameworks} are employed to simulate specific realizations 
of system dynamics. The primary tool for capturing bounded rationality, heterogeneity, and complex emergence from the bottom up is Agent-Based Modeling (ABM), 
which simulates the granular, adaptive behavior of autonomous entities~\cite{epstein1996growing, farmer2024quantitative}. 
For a more rigorous physical approach, Kinetic Monte Carlo (KMC) algorithms solve the Master Equation numerically by generating exact stochastic trajectories via 
event-by-event sampling based on physical rates~\cite{gillespie1976general, voter2007introduction, bottcher2020unifying}. 
Furthermore, Directed Percolation provides a geometric simulation approach to identify critical tipping points and connectivity thresholds in 
non-equilibrium spreading processes, distinguishing between local containment and systemic contagion~\cite{stauffer2018introduction, hinrichsen2000non}.
Finally, \textit{deterministic and aggregate frameworks} offer macroscopic approximations by simplifying the system to ignore fluctuations and correlations, 
describing dynamics in the limit of infinite system size. Evolutionary Population Dynamics, encompassing Generalized Lotka-Volterra equations and Replicator Dynamics, 
model the non-linear competition, selection, and cyclic dominance between aggregate populations driven by fitness differences~\cite{hofbauer1998evolutionary, nowak2006evolutionary}. 
Similarly, Mean-Field ODE Approximations are derived by averaging the Master Equation to describe the time-evolution of system densities, assuming that every agent interacts with the certain  
averaged variables~\cite{barrat2008dynamical, pastorsatorras2001epidemic}.}
% In order to model such systems, different frameworks from non-equilibrium frameworks are needed, such as 
% mean-field approximations~\cite{Barrat2008Dynamical}, 
% kinetic Monte Carlo methods~\cite{KineticMonteCarlo}, 
% percolation methods~\cite{Percolation}, 
% Self-Exciting Point Processes~\cite{bouchaud2024self}, 
% or Agent-Based Models~\cite{Farmer2024}.

\changed{Agent-based models (ABMs) offer an alternative simulational framework, they are computational systems that simulate the interactions of 
autonomous, heterogeneous entities to analyze how complex aggregate phenomena emerge from individual behaviors 
across social, biological, and technical domains \cite{epstein1996growing, grimm2006standard, bonabeau2002agent}.
 Within this field, Quantitative Agent-Based Models (QABMs) were recently introduced as a significant methodological advance, 
 utilizing bounded rationality and granular micro-data calibration to generate specific time-series forecasts. 
 This recent development offers a distinct alternative to traditional macroeconomic frameworks that rely on equilibrium assumptions 
 and rational expectations \cite{farmer2024quantitative}.}

\changed{In this paper, we will introduce abm-inspired neural networks that use the neural ODE framework with 
inductive biases from agent-based models to model out-of-equilibrium dynamics of complex systems.
Central inductive bias we use is the separation of laws into self-dynamics neural network and restricted graph neural network. Graph neural networks (GNNs) make these relational inductive biases explicit and 
have been used as learnable physics engines for interaction-driven systems~\cite{Battaglia2018Relational,SanchezGonzalez2019GraphPhysics}.
Restricted graph neural networks are a special class of graph neural networks that are restricted to only use the immediate neighbors of a node to 
 compute the output~\cite{Vasiliauskaite2024Generalization}.
 Graph Neural Networks operate as learnable message passing systems, where the exchange of information between nodes/agents is governed by differentiable neural 
 functions rather than fixed analytical equations~\cite{gilmer2017neural}. Thus, a GNN enables the system to ``learn physics'': 
 instead of relying on a human derivation of how agents should influence each other, the network explicitly learns the non-linear interaction 
 rules that best predict the system's evolution directly from data.
 Finally, by analogy to HNNs for Neural ODEs, we introduce additional structure to neural differential equations to model different conservation laws of agent-based models.}

\changed{In summary, this work: 
(i) introduces ABM-informed Neural Networks (ABM-NNs) for modeling complex systems while preserving structural constraints;
(ii) validates the approach on a generalized GLV system, recovering exact parameters of analytical laws from simulated data;
(iii) demonstrates superior out-of-sample generalization compared to GCN, SAGE, and GraphGPS baselines on contagion dynamics; and
(iv) applies the framework to empirical macroeconomic data, enabling counterfactual policy analysis via simulation.}

%\section{Learning dynamics in complex systems}
\section*{Methods}

% \begin{figure}[!ht]
%     \centering
%     \includegraphics[width=\linewidth]{figs/mak_nn_neuralpsi_combined.pdf}
%     % Generated by: python make_gnvf_demo.py (kept as a note, file name omitted from caption)
%     \caption{Learning complex dynamics (mass-action kinetics; protein–protein interaction dynamics) with a feed-forward neural network versus a graph neural vector field~\cite{Vasiliauskaite2024Generalization}. The GNVF leverages network inductive biases to extrapolate vector fields more accurately.}
%     \label{fig:neural-approx-motivation}
% \end{figure}
\begin{table}[!t]
    \centering
    \resizebox{\linewidth}{!}{
    \begin{tabular}{l l}
    \hline
    \textbf{Symbol} & \textbf{Description} \\
    \hline
    $\mathbf{x}$ & State vector variable \\
    $\dot{\mathbf{x}}$ & Time derivative $\frac{d\mathbf{x}}{dt}$ \\
    $A$ & Graph adjacency matrix \\
    $\theta$ & Parameters of the neural networks (NN)\\
    $\phi_{\theta}(\mathbf{x})$ & Self-interaction (Feed-Forward NN) \\
    $\psi_{\theta}(\mathbf{x}, A)$ & Interaction-with-neighbors (Graph NN) \\
    $x_i,y_i$ & State variables for node $i$ \\
    $w(t) = [\mathbf{x}(t), \mathbf{y}(t)]$ & State variable for all nodes \\
    $F(\phi_1, \phi_2, \psi_1)$ & Functional that combines neural networks \\
    $\mathcal{L}$ & Loss function \\
    $\Omega(\phi_1, \phi_2, \psi_1)$ & Total regularization \\
    $\mathcal{R}(\phi_i)$ & Individual-based regularization \\
    \hline
    \end{tabular}
    }
    \caption{Summary of symbols and notation used in the Methods section.}
    \label{tab:notation}
\end{table}

The structure of many complex systems is represented by a complex network (graph). When modeling the dynamics of complex systems, 
agent interactions are constrained by the underlying network. 
We first revisit the main results of the Graph Neural Vector Field framework~\cite{Vasiliauskaite2024Generalization}, 
since our approach extends this framework to agent-based models with multiple states.
%\subsection*{Single state dynamics on complex networks}
Consider a network of $N$ nodes with states $x_i \in \mathbb{R}^d$.  
In the graph neural vector field (GNVF) framework~\cite{Vasiliauskaite2024Generalization}, the dynamics is decomposed into self-interaction and 
interaction-with-neighbors components, both of which are parameterized by neural networks:
\begin{equation}
    \dot{\textbf{x}} = \phi_{\theta}(\textbf{x}) 
    + \psi_{\theta}(\textbf{x}, A),
    \label{eq:gnvf}
\end{equation}
where $\phi_{\theta}$ is the self-interaction neural network implemented as a feed-forward neural network, 
and $\psi_{\theta}$ is the interaction-with-neighbors neural network implemented as a special restricted graph neural network.
\changed{Now we define these operators more explicitly. The self-interaction function is a vector-valued function that maps the 
system state $\mathbf{x} \in \mathbb{R}^N$ to $\mathbf{x}' \in \mathbb{R}^N$, whose $i$-th element is computed by some function $f(x_i)$, that does not depend on the state of other nodes. 
The local interaction function is a vector-valued function that maps the system state $\mathbf{x} \in \mathbb{R}^N$ and graph adjacency 
matrix $A \in \mathbb{R}^{N \times N}$ to $\mathbf{x}' \in \mathbb{R}^N$, whose $i$-th element is computed by some function $g(x_i, \mathcal{N}(i))$, 
where $\mathcal{N}(i)$ is the set of neighbors of node $i$, i.e., $\mathcal{N}(i) = \{j : A_{ij} \neq 0\}$. } 

Note that a restricted graph neural network is needed~\cite{Vasiliauskaite2024Generalization} to enforce the correct 
inductive biases for the vector field: within infinitesimal time, a node can interact only with its immediate neighbors.
\changed{A restricted graph neural network is a graph neural network that is restricted to only use the immediate neighbors of a node to 
compute the output (i.e. local interaction function). In this paper, we use the terms restricted graph neural network and Graph Neural Vector Field interchangeably.}
%In Fig.~\ref{fig:neural-approx-motivation}, we revisit the main benefit of using the Graph Neural Vector Field framework in contrast to a pure feed-forward neural network.
%Here we revisit the main benefit of using the Graph Neural Vector Field framework in contrast to a pure feed-forward neural network.
The Graph Neural Vector Field framework can better capture the vector field of dynamics out of sample than 
a pure feed-forward neural network or even state-of-the-art graph neural networks~\cite{Vasiliauskaite2024Generalization}, 
because it incorporates an appropriate set of inductive biases.
%It is clear that GNVF can better capture the vector field of the dynamics out of domain, since it uses an appropriate set of inductive biases.
However, how to proceed in the case of general dynamics with multiple states is not clear. 
\changed{By analogy with Hamiltonian Neural Networks for Neural ODEs, we propose agent-based-model informed neural networks (ABM-NNs) 
that model the dynamics of complex systems, by incorporating principles of agent-based models into the neural networks. 
These principles can be connected to the laws of conservation of mass, interaction between agents, or other principles of agent-based models.}

\subsection*{ABM-informed neural networks}
% Previous chapter showed why pure graph neural networks are not enough to model dynamics on complex systems. In particular, Graph Neural Network Field~\cite{Vasiliauskaite2024Generalization}, introduced several important adaptations like: (i) restricted graph neural network for interaction and (ii) self-interaction neural network, that jointly work together to model complex dynamics. In this chapter, we will extend set of inductive biases. By analogy to how Hamiltonian Neural Networks have introduced energy conservation property to Neural ODE, here we will introduce additional structure to neural differential equations. 
% However, it is important to denote that there is no single universal principle like Lagrangian or Hamiltonian in the agent-based-model universe. 

We start with a compartmental agent-based model, where every node in the network can be in one of three states: $x,y,z$.
\begin{equation}
    \dot{x}_i = f(x,y,z,A)
\end{equation}
\begin{equation}
    \dot{y}_i = g(x,y,z,A) 
\end{equation}
\begin{equation}
    \dot{z}_i = h(x,y,z,A)
\end{equation}
Without loss of generality, framework also work for more than three states.

%Let's proceed with the monotone dynamics of transitions between states: $x \rightarrow y \rightarrow z$.
We will use $k$ possible feed-forward neural networks to model the self-interactions of each agent: $\phi_1, ... \phi_k$, and $k$
graph neural networks~\cite{Wu2019ComprehensiveSurvey} to model the interaction between agents: $\psi_1, ..., \psi_k$.
\begin{equation}
    \dot{x}_i = F(\phi_1, ..., \phi_k, \psi_1, ...,\psi_k),
\end{equation}
\begin{equation}
    \dot{y}_i = G(\phi_1, ..., \phi_k, \psi_1, ...,\psi_k),
\end{equation}
\begin{equation}
    \dot{z}_i = H(\phi_1, ..., \phi_k, \psi_1, ...,\psi_k),
\end{equation}
In order to model the dissipative nature of the system, we add a time-dependent function $d(t)$ to the constraints:
\begin{equation}
    F(.)+G(.)+H(.)=d(t),
    \label{eq:dissipative-constraint}
\end{equation}
where $F, G, H$ are the functionals that combine the self-interaction and interaction neural networks into the transition dynamics,
and $.$ denotes the inputs $\phi_1, ... ,\phi_k$, $\psi_1, ... ,\psi_k$.
The constraint \eqref{eq:dissipative-constraint} can be hard or soft.
For the case of a non-dissipative system, we have $d(t)=0$, which can be represented as a hard constraint.
In the case of a soft constraint, we learn the $F, G, H$ functions such that $d(t) \to 0$. 
Finally, in the case of a dissipative system, where some quantity/energy is irreversibly lost 
(e.g., as heat, sound, or radiation to the environment), we learn $d(t) \neq 0$.

The learning loss is represented as the sum of the data loss and regularization terms:
\begin{equation}
    \mathcal{L} \;=\; \mathcal{L}_{\text{data}}\; +\; \Omega(\phi_1, ..., \phi_k, \psi_1, ..., \psi_k).
\end{equation}
where $\Omega(\phi_1, ..., \phi_k, \psi_1, ..., \psi_k)$ is the regularization term.
The prediction $\hat{w}(t)$ is obtained by numerically solving the learned system with the learned parameters:
\begin{equation}
    \hat{w}(t) = \hat{w}(0) + \int_{0}^{t} \xi(\hat{w}(\tau)) \, d\tau,
\end{equation}
where $\hat{w}(t) = [\textbf{x}(t), \textbf{y}(t), \textbf{z}(t)]$, and $\xi(\cdot)=(F(\cdot),G(\cdot),H(\cdot))$ is the function that computes the derivative of the trajectory.

\changed{The data loss can be represented as the time integral of the $L_p$ norm between the two trajectories:
\begin{equation}
    \mathcal{L}_{\text{data}} \;=\; \int_{0}^{T} \big\|\, w(t) - \hat{w}(t) \,\big\|_p \, dt,
    \label{eq:macro-loss}
\end{equation}
where $w(t) = [\textbf{x}(t), \textbf{y}(t), \textbf{z}(t)]$ is the ground truth trajectory, and $\hat{w}(t)$
 is the predicted trajectory. In this paper, we will only the $L_1$ norm.
 In high-dimensional spaces, the $L_1$ norm is often more robust than the $L_2$ norm because 
 $L_2$ distances concentrate (near–far distances become similar) and become dominated by a few large coordinates, 
 while the $L_1$ norm preserves contrast between points and is less sensitive to outliers, mitigating the curse of dimensionality~\cite{Aggarwal2001HighDimDistances}.
 But at the same time, since we are proposing a general framework, other $L_p$ norms could be more appropriate for different applications.
 }
Regularization terms are any constraints on the functionals e.g. initial value, boundary value, non-negativity, etc.
\begin{equation}
    \Omega(\phi_1, ..., \phi_k, \psi_1, ..., \psi_k) = \sum_{i=1}^{k} \left(\lambda_{\phi_i} \mathcal{R}(\phi_i)+ \lambda_{\psi_i} \mathcal{R}(\psi_i)\right),
\end{equation}
where $\mathcal{R}(f)$ is the regularization term for the $i$-th self-interaction and interaction functions.
% write expetation over some dataset for regularization terms, we sample dataset values and in expectation we want to be close to the ground truth values
The regularization term can be expressed as the expectation over some dataset (initial values, boundary values, etc.) that consists
of pairs of input and output values $f(x_i)=y_i$ for function $f$:
\begin{equation}
    \mathcal{R}(f) = \mathbb{E}_{(x_i, y_i) \in \mathcal{D}} \big[\,\big|f(x_i) - y_i \big|\,\Big].
\end{equation}

Learning functionals $F$, $G$, $H$ can be implemented in different ways. 
By analogy to Fourier, Kernel methods, we can represent $F$, $G$, $H$ as linear combinations of the learned functions $\phi_i$ and $\psi_i$:
\begin{equation}
    F(\phi_1, ..., \phi_k, \psi_1, ..., \psi_k) = \langle \textbf{c}_{f} \phi \rangle + \langle \textbf{c}_{f} \psi \rangle,
\end{equation}
where $\langle \cdot \rangle$ denotes the dot-product $ \langle \textbf{c}_{f} \phi \rangle = \sum_{i=1}^{k} c_{f,i} \cdot \phi_i$.
Similarly, we can represent $G=\langle \textbf{c}_{g} \phi \rangle + \langle \textbf{c}_{g} \psi \rangle$, and $H=\langle \textbf{c}_{h} \phi \rangle + \langle \textbf{c}_{h} \psi \rangle$.
Finally, we would like to stress, that different case-studies require slightly different inductive-biases in order to better capture agent-based behavour in different complex systems, see results sections for more details.
% Similarly, we can represent $G$ and $H$ as:
% \begin{equation}
%     G(\phi_1, ..., \phi_k, \psi_1, ..., \psi_k) = \langle \textbf{c}_{g} \phi \rangle + \langle \textbf{c}_{g} \psi \rangle,
% \end{equation}
% \begin{equation}
%     H(\phi_1, ..., \phi_k, \psi_1, ..., \psi_k) = \langle \textbf{c}_{h} \phi \rangle + \langle \textbf{c}_{h} \psi \rangle.
% \end{equation}

\section*{Results}
In the results section we showcase the proposed framework through three case studies, each relaxing more assumptions than the previous one. 
The first study revisits a generalized Lotka--Volterra system with analytical dynamics, where only a handful of parameters are calibrated. 
It adapts population dynamics to a fictional cosmological scenario featuring five factions competing for resources, with $F=\phi+\psi$ fixed analytically and 
finite resources or carrying capacity is enforced via a predefined adjacency matrix with non-zero diagonals. 
The second study focuses on an SIR contagion model on graphs. We begin with analytical $F,G,H$ and learned neural operators $\phi,\psi$, then benchmark against GCN~\cite{Kipf2017GCN}, 
SAGE~\cite{Hamilton2017GraphSAGE}, and GraphGPS Transformer~\cite{Rampasek2022RecipeGPS} baselines, and additionally test robustness to observational noise, 
new initial conditions, 
evaluate out-of-sample generalization on unseen graphs, and finally learn the linear functionals directly from data. 
Finally, third study targets latent macroeconomic dynamics of top 10 largest economies in the world, where empirical time series inform both the interaction matrix and the 
subsequent counterfactual simulations of the learned system. In the macroeconomic model, we put special emphasis on heterogeneity by using Feature-wise Linear Modulation adapters 
(FiLM)~\cite{perez2018film}.

\changed{It is important to note that the framework presented in the methods section incorporates several explicit and 
implicit inductive biases relevant to the design of agent-based models. We establish spatiotemporal constraints by 
using restricted Graph Neural Networks to prioritize local peer interactions over global knowledge, 
while leveraging the discretization step of Neural ODEs to model the speed of information processing.
To capture human-like limitations, we utilize gradient descent to naturally seek satisficing heuristics rather than global optima. We combine this computational constraint with FiLM-based heterogeneity, effectively replacing the uniform representative agent with a diverse population. These behavioral mechanisms are further grounded by strict conservation laws and an empirical prior, ensuring that the agents' evolution remains physically consistent and aligned with historical data rather than abstract theory. Nevertheless, we also demonstrate that the framework remains sufficiently expressive to capture complex dynamics in scenarios where we simulate ground-truth behavior.}

\subsection*{Case Study 1: Analytical Recovery (Cosmological GLV)}
\changed{We first take a well-studied analytical GLV model and apply it into the fictional three-body problem trilogy~\cite{Liu2008ThreeBody}, 
where distinct sociopolitical factions (the mainstream Planetary Defense Council, the militaristic Solar Fleet, the strategic Wallfacers,
 the traitorous ETO, 
and the opportunistic Escapists) compete for limited industrial capacity and scientific resources to either save or 
destroy humanity.
 Each faction $i$ is treated as a node with capacity $X_i(t)$, restricted potential $r_i$, and 
 interaction weights $a_{ij}$, yielding the additive self-plus-interaction decomposition. Data generation process is defined as following:}
\begin{equation}
    \frac{dX_i}{dt} \;=\; 
    \underbrace{X_i\big(r_i\,S_i(t) + \tau_i(t)\big)}_{\text{self dynamics}}
    + \underbrace{X_i \sum_{j=1}^{N} a_{ij} X_j}_{\text{interaction term}},
    \label{eq:trisolaris-glv}
\end{equation}
 \changed{where $S_i(t)$ encodes Sophon-imposed physics access (extraterrestrial supercomputers that lock down human scientific progress in the novel), $\tau_i(t)$ is an exogenous policy regime, and $a_{ij}$ captures sabotage, cooperation, or spillovers. 
 Crucially, non-zero diagonal elements $a_{ii} \neq 0$ imply finite resources, effectively enforcing a carrying capacity that prevents unbounded growth.}
\changed{In the first case study, we want to show how ABM-inspired differential equations can be calibrated, even if different exogenous interventions are being applied. 
Optimization is performed with gradient descent-type optimizer, over the computational graph, that is unfolded with system of ordinary differential equations.}

Our learning model is represented as the following equation:
\begin{equation}
    \dot{\textbf{x}} =  \phi_1(\textbf{x}) + \psi_2(\textbf{x}, A),
    \label{eq:trisolaris-gnvf}
\end{equation}
where $\phi_1(x_i)=r_i x_i$ and $\psi(\textbf{x}, A)= \sum_{j=1}^{N} a_{ij} x_i x_j$.
We can think of  $\phi_1$ as very simple differentiable network that has a single parameter $r_i$ and 
$\psi$ as the interaction function with no-free parameters. 
We start with this simple case, since it is very close to classical ABM model. 
But the calibration of parameters, will be done by using automatic differentiation framework, over the differentiable 
ODE solver. 

\changed{In this fictional story, different factions are competing for resources, and different exogenous effects are happening. 
One is through the use of multiplicative factor $S_i(t)$, and second one 
via the use of additive factor $\tau(t)$. Both of these factors are affecting the effective growth rate of different factions in the generalized Lotka--Volterra system (GLV).
We will now describe exogenous effects, that we will use in our GLV system.}

 Appendix Eq.~\eqref{eq:SI-trisolaris-s} lists the exact step function used in the simulations. Full numerical values for the exogenous forcing $\tau_i(t)$ for all factions are listed in Appendix Eq.~\eqref{eq:SI-trisolaris-tau}.

\changed{During training we expose the learner only to $t\in[0,50]$ via curriculum rollouts, keeping $a_{ij}$ fixed and sparse while optimizing the growth scalars $r_i$ that enter 
the analytical pairwise term $\phi_{\text{pair}}(x_i,x_j)=x_i x_j$.  Evaluation integrates the learned right-hand side to $t=250$ to stress extrapolation well 
beyond the curriculum horizon.  Appendix Fig.~\ref{fig:three-body-explicit} confirms that the explicit parameterization both reproduces the highlighted regime 
transitions (top panel) and recovers the intrinsic growth rates from random initialization (bottom panel), demonstrating that auto-differentiable calibration can 
recover a closed-form ABM from short rollouts.} %For the detailed training algorithm, see Appendix~\ref{sec:algo-glv}.

\changed{Having validated the framework on an analytical GLV, we now replace the self and interaction operators with neural parameterizations while preserving the 
conservation-style constraints introduced in the Methods section.}

\subsection*{Case Study 2: Contagion Dynamics}

\begin{figure*}[t]
    \centering
    \includegraphics[width=0.98\linewidth]{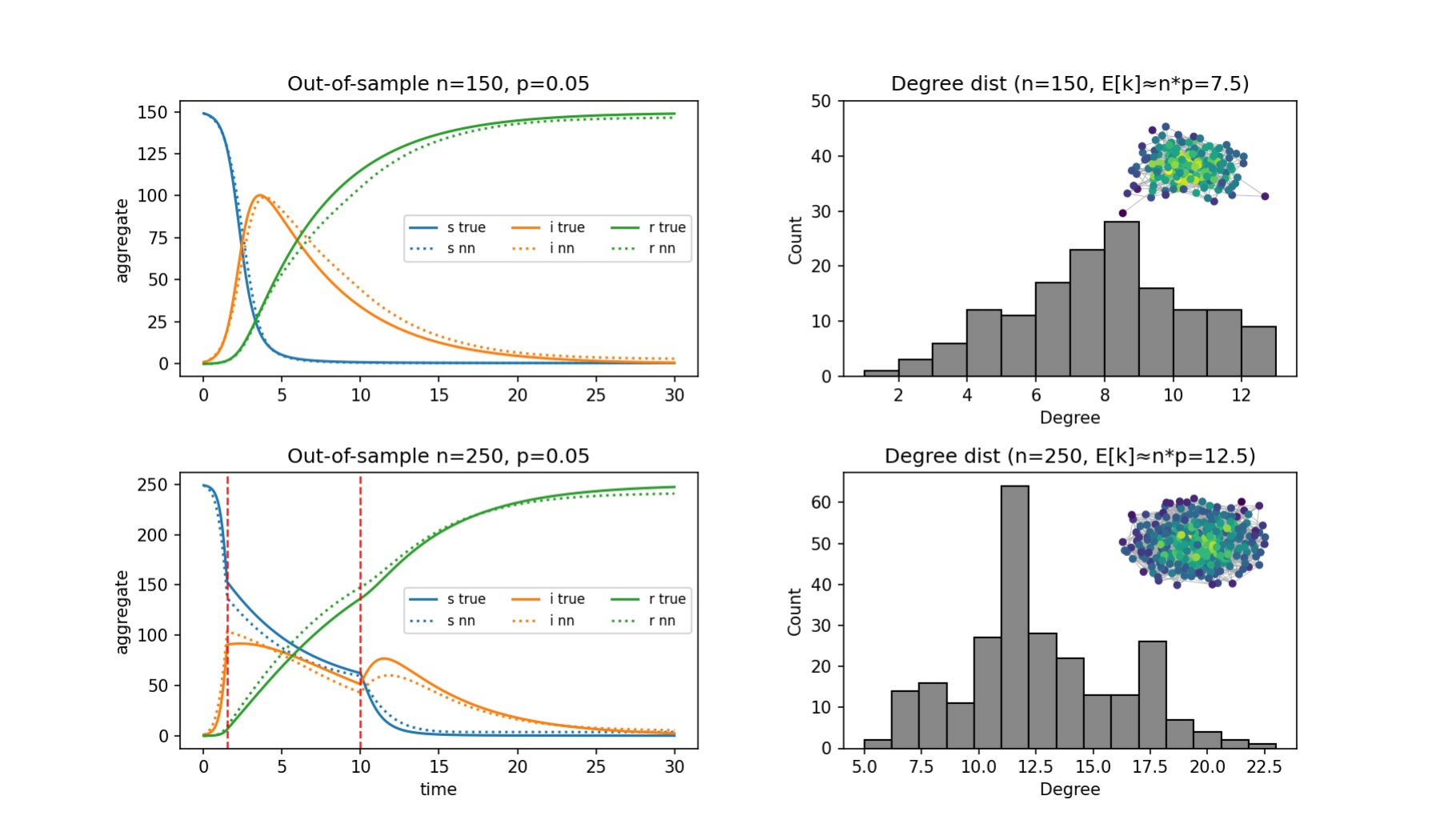}
    % Generated by: python epidemic_demo_macro_out_of_sample.py --load_model experiments/26_09_2025_17:09/macro_rhs.ckpt --nodes1 150 --p1 0.05 --nodes2 250 --p2 0.05 --beta 0.3 --gamma 0.2 --t_max 30.0 --dt 0.1 --seed 7 --restrict_time 1.5 --restrict_end_time 10.0 --restrict_drop_fraction 0.9
    \caption{Out-of-sample (OOS) application of a model learned from a single trajectory. Top row: deployment on a new Erd\H{o}s--R\'enyi graph (\(n=150\), \(p=0.05\)) without interventions. Bottom row: identical model evaluated on a larger graph (\(n=250\), \(p=0.05\)) under a social-distancing intervention that disables 90\% of links per node during the window \(t \in [1.5, 10.0]\) (marked by two red vertical lines). Despite not observing interventions during training, the model accurately predicts the system response with infection rate \(\beta=0.3\) and recovery rate \(\gamma=0.2\).}
    \label{fig:macro-sir-model-out-of-sample}
\end{figure*}

Next, we demonstrate the effectiveness of the proposed method on a prototypical example of a compartmental model, 
a well known SIR contagion dynamics~\cite{PastorSatorras2015Epidemic}. \changed{Data generation process is defined as following:}
\begin{equation}
    \dot{S}_i = -\sum_j \beta A_{i,j} S_i I_j
\end{equation}
\begin{equation}
    \dot{I}_i = \sum_j \beta A_{i,j} S_i I_j - \gamma I_i
\end{equation}
\begin{equation}
    \dot{R}_i = \gamma I_i
\end{equation}
This dynamics is an example of nonlinear, non-dissipative dynamics with multiple states that is out of equilibrium from a statistical 
mechanics perspective, and it exhibits critical phase transitions. 
Since dynamics is non-dissipative, we will use hard-constraint $d(t)=0$ to model the 
conservation of mass. We will use graph neural network to model the interaction between agents, 
and feed-forward neural network to model the self-interaction of each agent.

Since, we know the structure of SIR process, we can start with the hard-wired functionals $F, G, H$ for the ABM-informed neural 
network:
\begin{equation}
    F(\phi_1, \phi_2, \psi_1, \psi_2) = - \psi_1
\end{equation}
\begin{equation}
    G(\phi_1, \phi_2, \psi_1, \psi_2) = \psi_1 - \phi_2
\end{equation}
\begin{equation}
    H(\phi_1, \phi_2, \psi_1, \psi_2) = \phi_2
\end{equation}
The hard constraint $d(t)=0$ represents conservation of mass, since the sum of the states is constant.
Next, we model the graph neural network in the form of a graph neural vector field~\cite{Vasiliauskaite2024Generalization}:
 \begin{equation}
    \psi_1(.) = \sum_j A_{i,j}\,\phi_1(S_i, I_j)
\end{equation}

% By plugging the functions into the ABM-informed neural network, we obtain the following ABM-informed neural network for SIR dynamics:
% \begin{equation}
%     \dot{S}_i = -\sum_j A_{i,j}\,\phi_1(S_i, I_j)
% \end{equation}
% \begin{equation}
%     \dot{I}_i = \sum_j A_{i,j}\,\phi_1(S_i, I_j) - \phi_2(I_i)
% \end{equation}
% \begin{equation}
%     \dot{R}_i = \phi_2( I_i )
% \end{equation}
% which conserve mass per node by design $F+G+H=0$.
To learn a neural right-hand side from aggregate data only, we parameterize a pairwise transmission law
 \(\phi_1:\mathbb{R}^2\!\to\!\mathbb{R}_{\ge 0}\) taking local \(S_i\) and neighboring \(I_j\), 
 and a recovery law \(\phi_2:\mathbb{R}\!\to\!\mathbb{R}_{\ge 0}\) taking \(I_i\). 
 Both are implemented as small MLPs with LeakyReLU in hidden layers 
 and ReLU output to enforce nonnegativity and zero-at-zero behavior. 

The data-fitting loss is the time integral of the L1 norm between aggregated macroscopic state  
\(s(t)=\sum_i S_i(t)\), \(i(t)=\sum_i I_i(t)\), and \(r(t)=\sum_i R_i(t)\) and ground truth macro trajectories, using loss in Eq.~\eqref{eq:macro-loss}.
% \begin{equation}
% \mathcal{L}_{\mathrm{macro}} \;=\; \int_{0}^{T} \big\|\, w(t) - \hat{w}(t) \,\big\|_1 \, dt,
% \end{equation}
% which in practice is evaluated on the simulation grid (RK4 steps) as a Riemann sum. 

The regularization terms enforce the model to learn the following properties for the functions $\phi_1$ and $\phi_2$:
\begin{equation}
\mathcal{R}_{\phi_1,\mathrm{axis}} \;=\; \mathbb{E}_{S\sim\mathcal{U}[0,1]}\Big[\,\big|\phi_1(S,0)\big|\,\Big],
\end{equation}
\begin{equation}
\mathcal{R}_{\phi_2,\mathrm{origin}} \;=\; \big|\phi_2(0)\big|.
\end{equation}
We collect them into a single functional
\begin{equation}
\Omega(\phi_1,\phi_2) \;=\; \lambda_1\, \mathcal{R}_{\phi_1,\mathrm{axis}}\; +\; \lambda_2\, \mathcal{R}_{\phi_2,\mathrm{origin}}.
\end{equation}
The total objective is
\begin{equation}
\mathcal{L} \;=\; \mathcal{L}_{\mathrm{macro}}\; +\; \Omega(\phi_1,\phi_2).
\end{equation}
Regularization weights $\lambda_{\mathrm{axis}}=1$, $\lambda_{\mathrm{origin}}=1$ with a warmup of 100 epochs.
In Appendix Fig.~\ref{fig:macro-sir-training}, we show the training results for the ABM-informed neural network on SIR dynamics.
In this experiment (Appendix Fig.~\ref{fig:macro-sir-training}), we generate a single Erd\H{o}s--R\'enyi contact network with node 
count \(n=100\) and edge probability \(p=0.05\). 
Ground-truth macroscopic curves \(s(t)=\sum_i S_i(t)\), \(i(t)=\sum_i I_i(t)\), and \(r(t)=\sum_i R_i(t)\) 
are obtained by simulating an SIR model with infection and recovery rates \(\beta=0.4\) and \(\gamma=0.2\), 
respectively, using RK4 time stepping with \(\mathrm{d}t=0.1\) up to \(t_{\max}=60\). 
Initial conditions are set by infecting up to 50 nodes sampled from the largest connected component, 
with the remainder susceptible and no recoveries: \(S_i(0)\in\{0,1\}\), \(I_i(0)\in\{0,1\}\), \(R_i(0)=0\), and \(S_i(0)+I_i(0)=1\).
Training uses a differentiable RK4 rollout in PyTorch on the fixed graph and minimizes an integral L1 path loss between the 
stacked state trajectories with a simple curriculum on the rollout horizon: starting at \(t=1\) and increasing by
 \(\Delta t=1\) every 50 epochs up to a cap (default \(t_{\text{train}}\le 30\)). 
 We add a regularization functional \(\Omega(\phi_1,\phi_2)\) and optimize the sum of the data 
 loss and \(\Omega\). Appendix Fig.~\ref{fig:macro-sir-training} also shows the training losses and regularizers at different epochs. 
 We observe that the model learns the dynamics of the system, and the regularization terms enforce the desired properties.

After training, we simulate the learned ABM-informed neural network over the full horizon to compare 
\(\tilde{s}(t),\tilde{i}(t),\tilde{r}(t)\) with the ground truth, inspect 1D slices of \(\phi_1(0.5, x)\).
Importantly, the structure of the SIR equations preserves node-wise mass: for each \(i\), \(\dot S_i+\dot I_i+\dot R_i=0\). 
With simplex-consistent initial conditions \(S_i(0)+I_i(0)+R_i(0)=1\), the conservation law implies 
\(S_i(t)+I_i(t)+R_i(t)=1\) for all \(t\). 
Our architectural choices support this invariance, 
and the numerical scheme maintains it during simulation.
In Hamiltonian neural networks, the neural parameterization enforces a symplectic vector field, 
which preserves a first integral (the Hamiltonian/energy) by construction.
Analogously, in our epidemic setting the compartmental structure induces a conservation law (mass on each node). 
By designing the neural right-hand side so that \(\dot S_i+\dot I_i+\dot R_i=0\), we add inductive bias that is the mass-conserving 
counterpart of the energy-preserving bias in HNNs, ensuring that neural dynamics remain on the probability simplex for each agent.

Next, we turn our attention to the graph-based out-of-sample (OOS) application of the learned ABM-informed neural network.
In Fig.~\ref{fig:macro-sir-model-out-of-sample}, we show the OOS application of the learned ABM-informed neural network for SIR dynamics.
In this experiment, we generate a new Erd\H{o}s--R\'enyi contact network with different node count and edge probability.
We simulate the learned ABM-informed neural network over the full horizon to compare \(\tilde{s}(t),\tilde{i}(t),\tilde{r}(t)\) with 
the ground truth.
We observe that the model generalizes to the new graph structure and accurately reproduces the aggregate dynamics. 
In the bottom panel, we show the OOS application under a social-distancing intervention that 
is active only between the two red vertical lines (links are masked during this window).
We observe that the ABM-NN accurately predicts the system's response, despite never observing interventions during training.
Qualitatively, infection curves goes up prior to the intervention,
and rapidly down during when the intervention is active, 
and has a 2nd smaller peak after the intervention has ended.

\paragraph*{Baseline model comparison.}
\changed{Quantitative benchmarks against standard graph neural models substantiate these qualitative behaviors. We compare the performance of the ABM-NN with GCN~\cite{Kipf2017GCN}, SAGE~\cite{Hamilton2017GraphSAGE}, and Graph Transformer (GraphGPS)~\cite{Rampasek2022RecipeGPS}. Appendix Fig.~\ref{fig:baseline-comparison-gcn} visualizes one representative rollout versus a GCN, highlighting more accurate out-of-sample forecasts of the inductive-bias-aware model.}

\changed{All models are trained for 1000 epochs using the Adam optimizer with a learning rate of $10^{-4}$, minimizing the macroscopic $L_1$ loss derived from a differentiable RK4 rollout.
\textit{ABM-NN Architecture}: The ABM-NN decomposes dynamics into pairwise interactions ($\phi_1$) and self-recovery ($\phi_2$), both implemented as 3-layer MLPs (hidden dim: 32) with LeakyReLU activations.
\textit{Baseline Architectures}: The GCN baseline employs a stack of two \texttt{GCNConv} layers (hidden dimension 64) followed by a linear head. Similarly, GraphSAGE uses two \texttt{SAGEConv} layers (mean aggregation, hidden dimension 64) with a linear head. The GraphGPS transformer processes inputs through two GPS layers---each combining a local \texttt{GCNConv} with a 4-head global attention mechanism (hidden dimension 64)---before a final projection head.}

\changed{To stress-test generalization, we evaluate every model on Erd\H{o}s-R\'enyi graphs with $N\in\{50,100,150,200\}$ and $I_0\in\{1\%,5\%,10\%,20\%\}$, averaging over ten random trials. Table~\ref{tab:macro-table} reports the mean MAPE. ABM-NN consistently achieves the lowest error, remaining below $45\%$ except for the densest cases, while transformers are the strongest baseline and GCN/SAGE degrade rapidly as graph size grows.}

\begin{table}[!t]
    \centering
    \resizebox{\linewidth}{!}{
    \begin{tabular}{cccccc}
\hline
Network Size & $I_0$ (\%) & ABM-NN & GCN & SAGE & TRANSFORMER \\
\hline
50 & 1\% & 12.86\% & 65.01\% & 84.16\% & 66.79\% \\
50 & 5\% & 13.88\% & 52.44\% & 94.29\% & 62.18\% \\
50 & 10\% & 15.51\% & 47.31\% & 98.30\% & 60.94\% \\
50 & 20\% & 22.76\% & 41.84\% & 127.55\% & 71.92\% \\
100 & 1\% & 20.43\% & 38.39\% & 38.08\% & 37.26\% \\
100 & 5\% & 22.69\% & 43.33\% & 48.40\% & 42.35\% \\
100 & 10\% & 28.62\% & 47.28\% & 41.64\% & 42.23\% \\
100 & 20\% & 30.00\% & 47.43\% & 44.23\% & 42.57\% \\
150 & 1\% & 28.91\% & 61.53\% & 77.42\% & 56.02\% \\
150 & 5\% & 32.22\% & 72.36\% & 95.87\% & 65.25\% \\
150 & 10\% & 30.61\% & 69.55\% & 82.60\% & 62.18\% \\
150 & 20\% & 43.83\% & 77.15\% & 115.59\% & 66.55\% \\
200 & 1\% & 42.23\% & 136.97\% & 362.73\% & 119.99\% \\
200 & 5\% & 57.76\% & 200.41\% & 525.58\% & 161.52\% \\
200 & 10\% & 57.24\% & 200.89\% & 485.88\% & 157.22\% \\
200 & 20\% & 50.21\% & 219.77\% & 693.14\% & 162.49\% \\
\hline
Average & -- & 31.86\% & 88.85\% & 188.47\% & 79.84\% \\
\hline
\end{tabular}
    }
    \caption{Out-of-sample MAPE (\%) across graph sizes and initial infection ratios. Each row averages ten random trials; lower is better.}
    \label{tab:macro-table}
\end{table}

\changed{Finally, noise-stress experiments in Appendix Fig.~\ref{fig:SI-noise-robustness} show that ABM-NN continues to track latent dynamics even when the aggregate supervision is heavily corrupted after $t=3$. Notably, observational noise can act as a form of regularization that stabilizes training~\cite{Liu2019NeuralSDE,Ghosh2020STEER,Oganesyan2020StochasticNeuralODE}, helping the model generalize to unseen conditions while maintaining the learned structure-preserving dynamics.}

Beyond the neural-only SIR study we also tested a functional-learning variant in 
which $F$ is fixed for interpretability while $G$ and $H$ are learned from data. 
To keep those functional coefficients well behaved we decouple the optimizers:
 the neural vector field ($\phi_1,\phi_2$) uses a smaller learning rate, whereas $G$ and $H$ receive a 
 larger rate and remain subject to the conservation penalty. 
 The resulting training remains stable under curriculum rollout and gradient clipping, 
 and the learned functional terms continue to respect mass balance despite observing only macroscopic losses.
  Full optimizer settings, traces, and trajectories for this configuration are documented in 
  Appendix Fig.~\ref{fig:SI-separate-lr-training}.

\subsection*{Case Study 3: Macroeconomic Dynamics}

\changed{The final case study couples GLV structure with a full neural ODE for latent macro channels, so both the interaction network and macro feedbacks are learned directly from data i.e. 
data generation process that drives the underlying global world economy is not known in this case.
The learning model, consists of two parts: a GLV-based neural network and a neural ODE for latent macro channels.
Each country $i$ carries a normalized GDP state $x_i(t)$ (GDP divided by its 1995 value) and a 
latent macro vector $y_i(t)\in\mathbb{R}^6$ that summarizes the macroeconomic state of the country.
We retain the self/interaction decomposition}
% \begin{equation}
%     \dot{\textbf{x}} = \underbrace{\phi_{1}(x_i,y_i,e_i)}_\text{self-interaction} + \underbrace{\sum_j A_{ij}\,\phi_{2}(x_i,x_j,e_i,e_j)}_\text{interaction term},
%     \label{eq:gnvf-macro-x}
% \end{equation}
\begin{equation}
    \dot{\textbf{x}} = \underbrace{\phi_{x}(x_i,y_i,e_i)}_\text{self-interaction} + \underbrace{\psi(\textbf{x}, \textbf{A}, e_i,e_j)}_\text{interaction term},
    \label{eq:gnvf-macro-x}
\end{equation}
\begin{equation}
    \dot{\textbf{y}} = \underbrace{\phi_{y}(y_i,x_i,e_i)}_\text{self-interaction},
    \label{eq:gnvf-macro-y}
\end{equation}
\changed{where $\textbf{x}_i$ is gross domestic product (GDP) of country $i$, $\textbf{y}_i = [\textbf{y}_i^1, ..., \textbf{y}_i^6]^T$,  with the following components: $\textbf{y}_i^1$ is unemployment, $\textbf{y}_i^2$ is interest rate, 
$\textbf{y}_i^3$ is debt, $\textbf{y}_i^4$ is working age population, $\textbf{y}_i^5$ is inflation, and $\textbf{y}_i^6$ is current account, and $e_i$ is the static embedding for country $i$.
The interaction term is a graph neural vector field, which is defined as:}
\begin{equation}
\psi(x_i,\textbf{x}, \textbf{A}, e_i,e_j)=\sum_j A_{ij}\,\phi_{2}(x_i,x_j,e_i,e_j), 
\label{eq:gnvf-macro-psi-1}
\end{equation}
where $\phi_2$ is a feed-forward neural network, and $\textbf{A}$ is the adjacency matrix of the graph that 
is learned from the data.

\paragraph*{Neural parameterization and adapters.}  
\changed{The shared self map $\phi_x$ follows the decomposition of Eq.~\eqref{eq:phi-decomp},}
\begin{equation}
    \phi_{1}(x_i,y_i,e_i) = x_i\,\phi_{g}(x_i,y_i,e_i) + \phi_{d}(x_i,y_i,e_i),
    \label{eq:phi-decomp}
\end{equation}
\changed{where $\phi_g(\cdot)$ is a feed-forward neural network that models annualized growth and $\phi_d(\cdot)$ is a feed-forward neural network that captures structural drifts. 
The latent macro network $\phi_{3}$ is a feed-forward neural network that models the dynamics of the macro channels.
For economic modeling and macroeconomic data scarcity, we would like to have parsimonious models, 
we will use a shared feed-forward neural network for both 
$\phi_x(\cdot)$ and $\phi_y(\cdot)$ and $\psi$.
Shared models correspond to the idea that economic laws follow some hidden economic universality laws or principles, and the same economic laws apply to all countries.
However, in reality, each country has additional hidden variables or factors that are not captured by the shared model, 
and thus we will use FiLM-style adapters to model the country-specific conditional dynamics.
Feature-wise Linear Modulation adapters (FiLM) were originally proposed for visual reasoning~\cite{perez2018film}, and they have been shown to be effective in learning conditional models.
In our case, we will use FiLM-style adapters to model the country-specific conditional macro dynamics. }

\begin{figure*}[!t]
    \centering
    % Experiment: experiments/GLV_learning_micro_gdp_extended_out_run_2025_11_25_23_43_07 (shared-parameter universal GLV)
    % Command:
    % /Users/nino-aisot/CodingProjects/abm-nn/.venv/bin/python GLV_learning_micro_gdp_extended_universal_out.py --device cpu --start-year 1995 --end-year 2024 --top-n 10 --train-cut-year 2021 --epochs 500 --steps 1 --lr 5e-4 --cyclical-max-lr 2e-3 --cyclical-step 75 --weight-decay 1e-4 --grad-clip 2.0 --curriculum-start 5 --curriculum-increment 4 --curriculum-period 10 --stagnation-epochs 50 --hidden-dim 8 --num-hidden 2 --phi-normalize-input --pretrain-epochs 500 --pretrain-batch-size 256 --pretrain-lr 1e-3 --relative-eps 1e-2 --interaction-warmup-epochs 25 --interaction-ramp-epochs 50 --lambda-a 1e-5 --a-max-abs 0.6 --interaction-gain 1.0 --exp-prefix GLV_learning_micro_gdp_extended_out_run --preview-interval 50
    \includegraphics[width=0.65\linewidth]{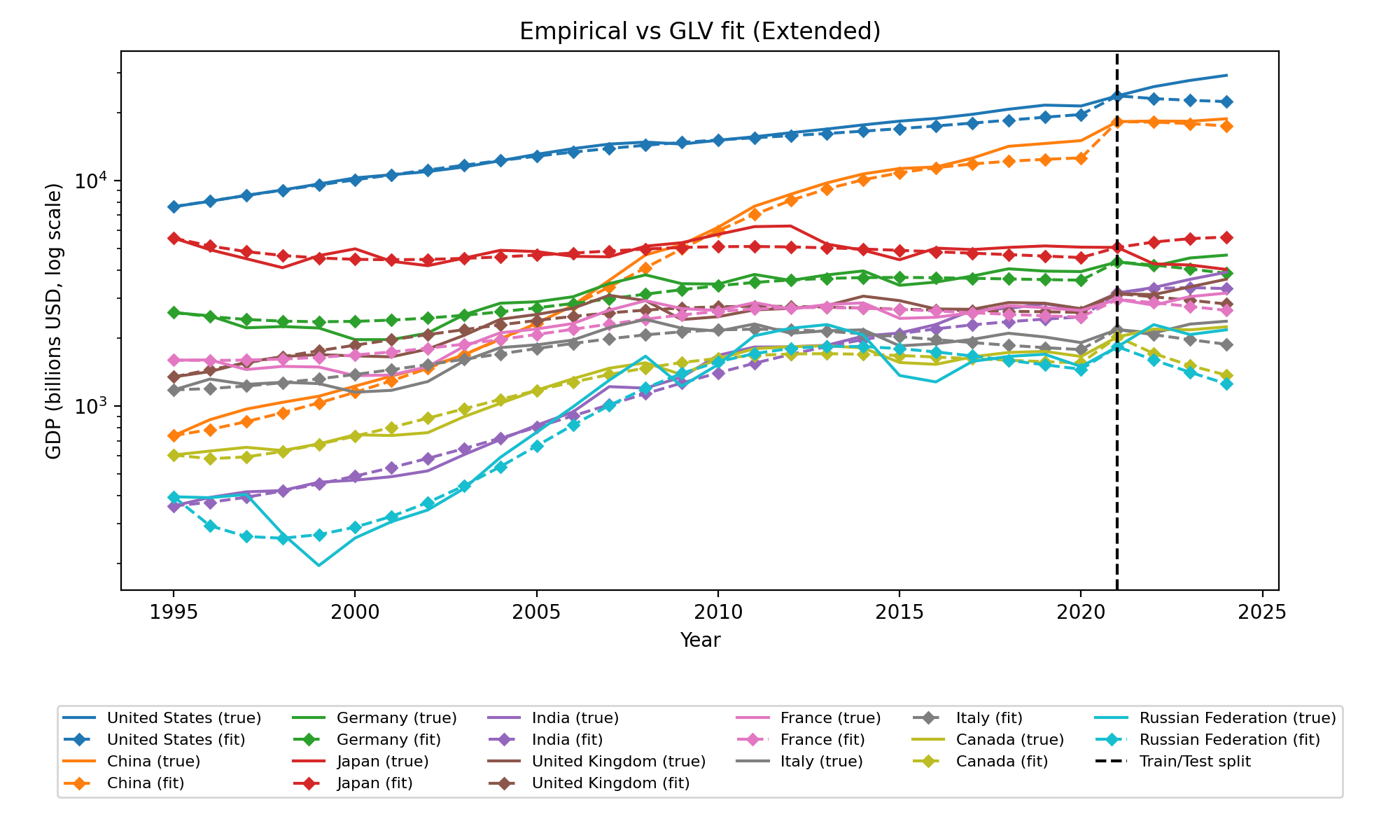}
    \vspace{3mm}
    \includegraphics[width=0.65\linewidth]{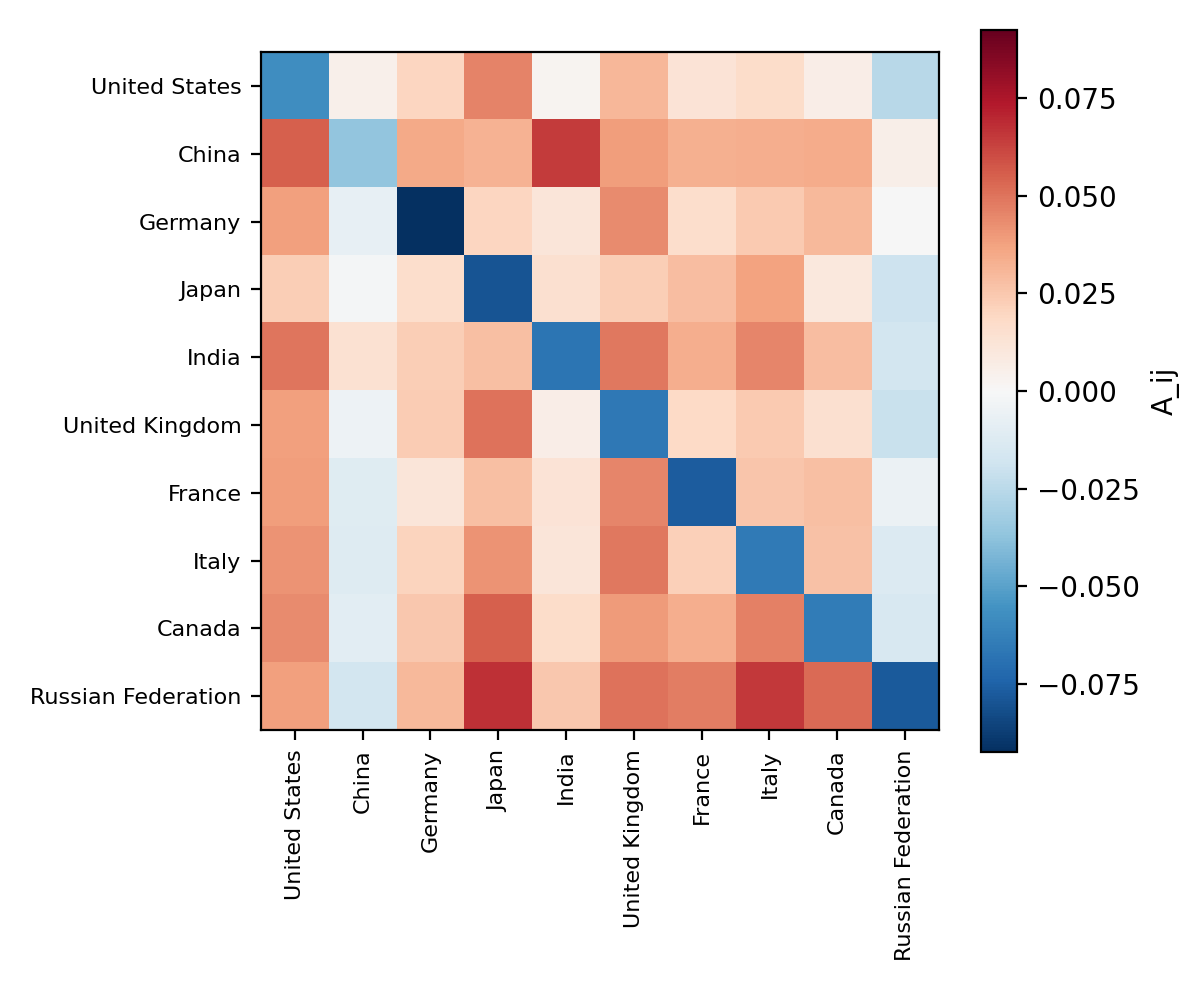}
    \vspace{2mm}
    \caption{\changed{Shared-parameter GLV trained on 1995--2020 GDP and macro data, evaluated out-of-sample through 2024 (top) together with the learned interaction matrix $A$ (bottom). The dashed line marks the 2021 holdout boundary. Full hyperparameters and training configuration are detailed in Appendix~\ref{sec:macro_hyperparams}.}}
    \label{fig:macro-traj-interaction}
\end{figure*}

\changed{FiLM-style adapters operate on each hidden layer $\ell$ of the shared MLP by conditioning on the country-specific embedding $e_i$ as follows:}
\begin{equation}
    h_{\ell+1} = \gamma_\ell(e_i) \odot h_\ell + \beta_\ell(e_i), \qquad \gamma_\ell,\beta_\ell : \mathbb{R}^{d_e} \to \mathbb{R}^{d_\ell},
    \label{eq:film}
\end{equation}
\changed{where $h_\ell \in \mathbb{R}^{d_\ell}$ denotes the activations at layer $\ell$, $e_i \in \mathbb{R}^{d_e}$ is the static embedding for country $i$, and $\odot$ is element-wise multiplication.  
Intuitively, the adapters rescale and shift each hidden unit so that heterogeneous countries can share the same base network while having country-specific gain and bias patterns; 
this greatly reduces the parameter count relative to training a separate network per economy, yet retains the ability to express qualitatively different responses to shocks.  
The latent macro network $\phi_{3}$ receives $[y_i,x_i,e_i]$ and outputs $\dot{y}_i$, letting unemployment, inflation, and other channels react causally to GDP and other macro channels deviations 
within the same differentiable rollout. For more details on the mathematical formulation for the forward pass of the macro-economic model, detailing the element-wise dynamics for country $i$ and 
the FiLM adaptation mechanism, see Appendix~\ref{sec:appendix-macroeconomic-case-study}.}

\paragraph*{State construction and causal imputation.}  \changed{The six components of $y_i$ summarize measured macro channels available in the World Bank~\footnote{\protect\url{https://data.worldbank.org/}} and Federal Reserve Economic DataFRED~\footnote{\protect\url{https://fred.stlouisfed.org/}} data sources.  
Unemployment (share of the labor force actively seeking work) captures slack in domestic demand.  
The lending interest rate on domestic credit approximates the marginal borrowing cost faced by firms and households, signaling the tightness of financial conditions.  
Central-government debt as a percentage of GDP tracks sovereign leverage and the scope for fiscal support.  
The working-age population share (15--64) acts as a demographic supply proxy tied to potential output and savings behavior.  
CPI inflation (year-over-year) records price-level dynamics that shape real interest rates and monetary policy reactions. 
Finally, the current-account balance as a share of GDP aggregates net trade plus income flows to indicate whether growth is financed internally or relies on external demand.}
\changed{Before standardization we fill missing entries with a strictly causal Kalman filter that assumes constant velocity and only conditions on past observations, thereby preventing leakage from future shocks.  After imputation each series is standardized per country.  Appendix Fig.~\ref{fig:SI-macro-covariates} summarizes the resulting macro trajectories for the ten largest economies.}

\paragraph*{Interaction structure and stabilization.}  \changed{The interaction term $\psi$ reduces to the classical GLV form $x_i(\max(x_j,0)+\varepsilon)^{\beta}$ when 
adapters are neutral.  We learn $\beta$ jointly with $A$ (converging to $0.2$--$0.4$ in practice) to attenuate spillovers from very large economies,
 and optionally gate each row with $\tanh(\omega_i)$ when interaction scaling is enabled.  The matrix $A$ starts from a weakly coupled baseline ($A_{ij}=1/N$, $A_{ii}=0$).}

\paragraph*{Macro results and interventions.}  \changed{Training on 1995--2021 data followed by 2021--2024 evaluation shows that the 
shared-parameter GLV reproduces the GDP trajectories in-sample, while learning interpretable couplings, as illustrated in Fig.~\ref{fig:macro-traj-interaction}. 
Note, that the inferred interaction matrix $A$ for GLV model is not symmetric e.g. China profits more from US GDP than US profits from China GDP, 
while US and Japan are in mutualism relationship. }

\changed{We implement counterfactual edits such as zeroing the bilateral US--China 
channels $A_{\text{US},\text{CN}}$ and $A_{\text{CN},\text{US}}$, and analyze the impact on the global economy.
In the appendix Fig.~\ref{fig:intervention-simulation}, we show the simulation until 2028 with baseline forecasts, and the edited intervention rollout initialized from the true 2024 state, 
highlighting how severing a single pair of links modifies the GDP trajectories for the US and China, and the global economy. One can observe that all global economies go down, 
but the largest effect is on the US and China. Note, that the roll-out of 4-year ahead is only used to illustrate the counterfactual effect, see Section~\ref{sec:policy-implications-and-ethics-statement} for 
the policy implications and ethics statement.}

\section*{Conclusion}
% pros and cons of ABMs and NNs
Agent-based models (ABMs) and neural networks (NNs) each offer distinct strengths and limitations. 
ABMs are powerful for capturing heterogeneity, local interactions, and emergent dynamics, making 
them well-suited for exploring complex adaptive systems, but they can be computationally expensive, 
hard to calibrate or validate, and difficult to interpret due to their complexity. 
Neural networks, on the other hand, excel at learning nonlinear patterns from large datasets 
and can approximate highly complex functions, yet they require substantial data and computational 
resources, risk overfitting, and function largely as black boxes. 
Moreover, unlike ABMs, NNs typically capture correlations rather than causality and thus 
cannot directly support counterfactual analysis without additional causal frameworks. 
Together, these contrasts highlight that the choice between ABMs and NNs depends on whether 
the priority lies in modeling mechanisms and emergent behavior or in predictive accuracy and 
pattern recognition.

We presented ABM-inspired neural networks that embed key principles of agent-based modeling into neural differential equations. 
The proposed formulation decomposes node-wise dynamics into distinct neural networks for self-interaction and interaction with neighbors, 
preserves per-node mass by construction, and can be trained using only a single macroscopic trajectory via an integral $L_1$ path 
loss augmented with lightweight regularizers.

Empirically, we showed that a model trained on a single trajectory generalizes out-of-sample to unseen graphs, 
accurately reproducing aggregate epidemic curves. 
We further demonstrated intervention analysis: despite never observing interventions during training, 
the learned model reliably predicts the system's response, with predicted curves closely matching the ground truth.
In particular, we showed that the model accurately captures the response to a time-bounded social-distancing window, 
highlighting its utility for counterfactual scenarios and control~\cite{Bottcher2022AIPontryagin}.
The key reason these models succeed in counterfactual analysis is that they learn the underlying dynamics of the system 
rather than mere correlations, enabling them to predict responses to novel interventions.
%However, the model cannot respond to arbitrary interventions; it is limited to changes in the interaction structure or additive terms in the differential equations of the form $\dot{x}=F(\cdot)+\delta(\cdot)$, where $\delta(\cdot)$ represents the intervention or control action. 
\changed{However, the model cannot respond to arbitrary interventions; it is limited to the interventions that can be incorporated 
into the learned differential equations,
via additive or multiplicative terms in the differential equations, and is not fundamentally changing the underlying dynamics.
We have compared the model to other graph neural networks in neural ODE setting, and showed that abm-informed neural networks can better capture 
the dynamics out of sample than the state-of-the-art graph neural networks. 
Furthermore, abm-nn model is robust to small perturbations in the data, such as observational noise, and even, 
it can act as a regularization term for the stability of the model.
Finally, we have applied the model to the macroeconomic case study, and showed that it can learn interpretable couplings between macro variables and GDP, 
and can be used to simulate counterfactual scenarios by editing the interaction structure.
}\\

\changed{By moving away from the theoretical purity of perfect rationality and equilibrium assumptions, our framework better reflects the reality of complex systems, where agents are not perfectly rational and the system is not operating in equilibrium. We enforce informational boundedness via restricted Graph Neural Networks, ensuring agents react only to local peers rather than global aggregates. Furthermore, the Neural ODE framework allows us to model bounded rationality via several mechanisms: solver step size, optimizer properties, and the complexity of the neural network architecture.}

\changed{The discretization step of the underlying solver acts as a proxy for cognitive latency, where the step size dictates how quickly agents can process and react to messages from their neighbors. This temporal constraint is matched by a computational one; the use of gradient descent calibrated against empirical data implies that agents learn 'good enough' local optima—heuristics—rather than global solutions, mirroring human satisficing behavior.}

\changed{While our framework allows each agent to possess unique decision-making logic (distinct weights)—an approach that works well in data-rich, simulated environments—real-world economic scenarios are often data-scarce. Therefore, we prioritize model parsimony by employing a shared feed-forward neural network modulated by Feature-wise Linear Modulation (FiLM) adapters for each agent.}

\changed{By combining this FiLM-based heterogeneity with restricted graph topologies, we effectively replace the frictionless 'representative agent' with a population of diverse, friction-constrained actors. Ultimately, by integrating these temporal, computational limits with heterogeneity, conservation laws, the model offers a plausible mechanism for simulating how constrained agents navigate structural changes.}

\changed{To conclude, we believe that the proposed framework is not a final solution, but rather a starting point 
for principled, interpretable neural modeling of complex systems. 
Future work will explore stochastic differential equation (SDE) formulations to explicitly model process noise and 
uncertainty quantification, extended Kalman filtering applications, online learning of the dynamical rules, 
as well as scaling to larger heterogeneous agent populations for more realistic modelling.}

\section*{Abbreviations}
\noindent
\textbf{ABM}: Agent-Based Model / Agent-Based Modeling \\
\textbf{ABM-NN}: Agent-Based-Model informed Neural Network \\
\textbf{CPI}: Consumer Price Index \\
\textbf{ETO}: Earth-Trisolaris Organization (fictional~\cite{Liu2008ThreeBody}) \\
\textbf{FFN}: Feed-Forward Network \\
\textbf{FiLM}: Feature-wise Linear Modulation \\
\textbf{FRED}: Federal Reserve Economic Data \\
\textbf{GCN}: Graph Convolutional Network \\
\textbf{GDP}: Gross Domestic Product \\
\textbf{GLV}: Generalized Lotka-Volterra \\
\textbf{GNN}: Graph Neural Network \\
\textbf{GNVF}: Graph Neural Vector Field \\
\textbf{GraphGPS}: Graph General Powerful Scalable Transformer \\
\textbf{GraphSAGE}: Graph Sample and Aggregate \\
\textbf{HNN}: Hamiltonian Neural Network \\
\textbf{KMC}: Kinetic Monte Carlo \\
\textbf{MAPE}: Mean Absolute Percentage Error \\
\textbf{MLP}: Multi-Layer Perceptron \\
\textbf{NN}: Neural Network \\
\textbf{ODE}: Ordinary Differential Equation \\
\textbf{OOS}: Out-of-sample \\
\textbf{PDE}: Partial Differential Equation \\
\textbf{QABM}: Quantitative Agent-Based Model \\
\textbf{ReLU}: Rectified Linear Unit \\
\textbf{RNN}: Recurrent Neural Network \\
\textbf{SDE}: Stochastic Differential Equation \\
\textbf{SIR}: Susceptible-Infected-Recovered

% In a follow-up study, we will apply the approach to other dynamical domains, including multi-agent systems for world trade 
% and artificial financial markets with heterogeneous agents (noisy traders and fundamental traders) to replicate 
% stylized facts of financial markets. Furthermore, we will extend the approach to incorporate 
% mean-field models~\cite{PastorSatorras2015Epidemic} at the degree-based level, enabling scalability to larger graphs and system sizes.\\

\section*{Policy implications and Ethics Statement}
\label{sec:policy-implications-and-ethics-statement}
The models presented in this paper fall within the category of agent-based models coupled with neural networks. 
If these models are used to inform policy recommendations, they should be evaluated with extra due diligence, 
where all assumptions are explicitly reported together with their limitations. 
Models that are evaluated only in simulation environments have inherent limitations that arise from 
the fidelity and scope of the simulation environment itself. Real-world deployment requires 
careful validation against empirical data and consideration of factors not captured in the simulation.
Direct interpretation of Generalized Lotka-Volterra macroeconomic model should be done with caution, 
as it is a simplified model of the real world dynamics, and implications can have serious consequences, 
especially in the case of global economic tensions.
Finally, note, that real-world systems are self-adaptive, and it may not be trivial 
to separate the effect of the intervention to learned dynamics from the self-adaptive response adaptation
of the dynamical rules $\phi_x, \phi_y, \psi \rightarrow \phi_x', \phi_y', \psi'$.
For more realistic applications, one should assume that the dynamical rules are adapting in an online manner.

Any model of socio-techno-economic systems, should not be used in 
unethical ways to harm people, countries, or organizations. Practitioners deploying these models 
must consider potential misuse, including discriminatory applications, manipulation of social systems, 
or deployment without informed consent of affected populations.
Models of social systems should be designed with the principle in mind that they increase the well-being of humans. 
Decision-makers must ensure that model-driven interventions respect democratic principles, 
protect human rights, and account for distributional equity and justice.

\section*{Declarations}
\vspace{-0.5em}
\subsection*{Availability of data and material}
\vspace{-0.5em}
The code used to generate the results will be available at \url{https://github.com/ninoaf}.
\subsection*{Competing interests}
\vspace{-0.5em}
The author has no competing interests to declare.
\vspace{-1em}
\subsection*{Funding}
\vspace{-1em}
Not available.
\vspace{-1em}
\subsection*{Authors' contributions}
\vspace{-1em}
N.A.F. designed research, performed research, wrote the paper.
\vspace{-1em}

% \begin{table}[h]
%     \centering
%     \small
%     \begin{tabular}{l|l}
%     \textbf{Abbreviation} & \textbf{Meaning} \\
%     \hline
%     ABM & Agent-Based Model \\
%     ABM-NN & Agent-Based-Model informed Neural Network \\
%     CPI & Consumer Price Index \\
%     ETO & Earth-Trisolaris Organization-fictional~\cite{Liu2008ThreeBody} \\
%     FFN & Feed-Forward Network \\
%     FiLM & Feature-wise Linear Modulation \\
%     FRED & Federal Reserve Economic Data \\
%     GCN & Graph Convolutional Network \\
%     GDP & Gross Domestic Product \\
%     GLV & Generalized Lotka-Volterra \\
%     GNN & Graph Neural Network \\
%     GNVF & Graph Neural Vector Field \\
%     GraphGPS & Graph General Powerful Scalable Transformer \\
%     GraphSAGE & Graph Sample and Aggregate \\
%     HNN & Hamiltonian Neural Network \\
%     KMC & Kinetic Monte Carlo \\
%     MAPE & Mean Absolute Percentage Error \\
%     MLP & Multi-Layer Perceptron \\
%     NN & Neural Network \\
%     ODE & Ordinary Differential Equation \\
%     OOS & Out-of-sample \\
%     PDE & Partial Differential Equation \\
%     QABM & Quantitative Agent-Based Model \\
%     ReLU & Rectified Linear Unit \\
%     RNN & Recurrent Neural Network \\
%     SDE & Stochastic Differential Equation \\
%     SIR & Susceptible-Infected-Recovered \\
%     \end{tabular}
%     \caption{Abbreviations and acronyms used in the manuscript.}
%     \label{tab:abbreviations}
%     \end{table}

\section*{Acknowledgements}
We would like to thank Prof. Helbing for organization of the workshop "Smarter Cities and Societies.
Back to the Future. What We Can and Cannot Optimize For." at ETH Zurich from 10-14 February 2025.
Workshop has sparked the interest for writing this paper.\\

\bibliographystyle{apsrev4-2}
\bibliography{refs}

\clearpage
\appendix
\section*{Appendix}

\section*{Appendix A: Cosmological social science case study}
\label{sec:appendix-GLB-trisolaris}

\begin{figure*}[!t]
    \centering
    % Experiment: experiments/GLV_learning_micro_2025_11_26_16:19:07
    % Command: /Users/nino-aisot/CodingProjects/abm-nn/.venv/bin/python GLV_learning_micro_explicit.py --epochs 500 --lr 5e-4 --cyclical-max-lr 2e-2 --cyclical-step 75 --t-max 250 --train-end 50 --curriculum-start 5 --curriculum-increment 4 --curriculum-period 50 --grad-clip 2.0 --stagnation-epochs 50
    \includegraphics[width=0.7\linewidth]{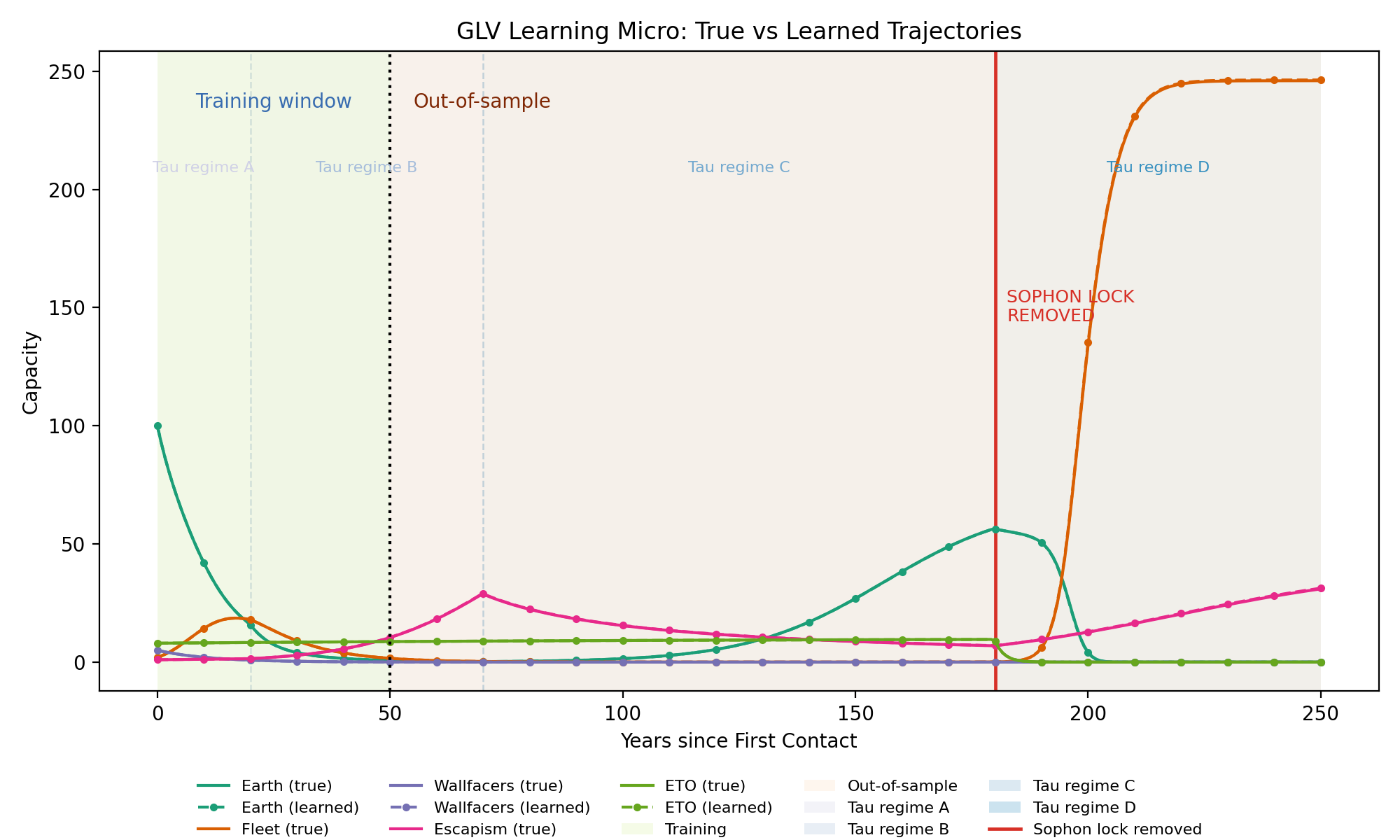}
    \vspace{2mm}
    \includegraphics[width=0.7\linewidth]{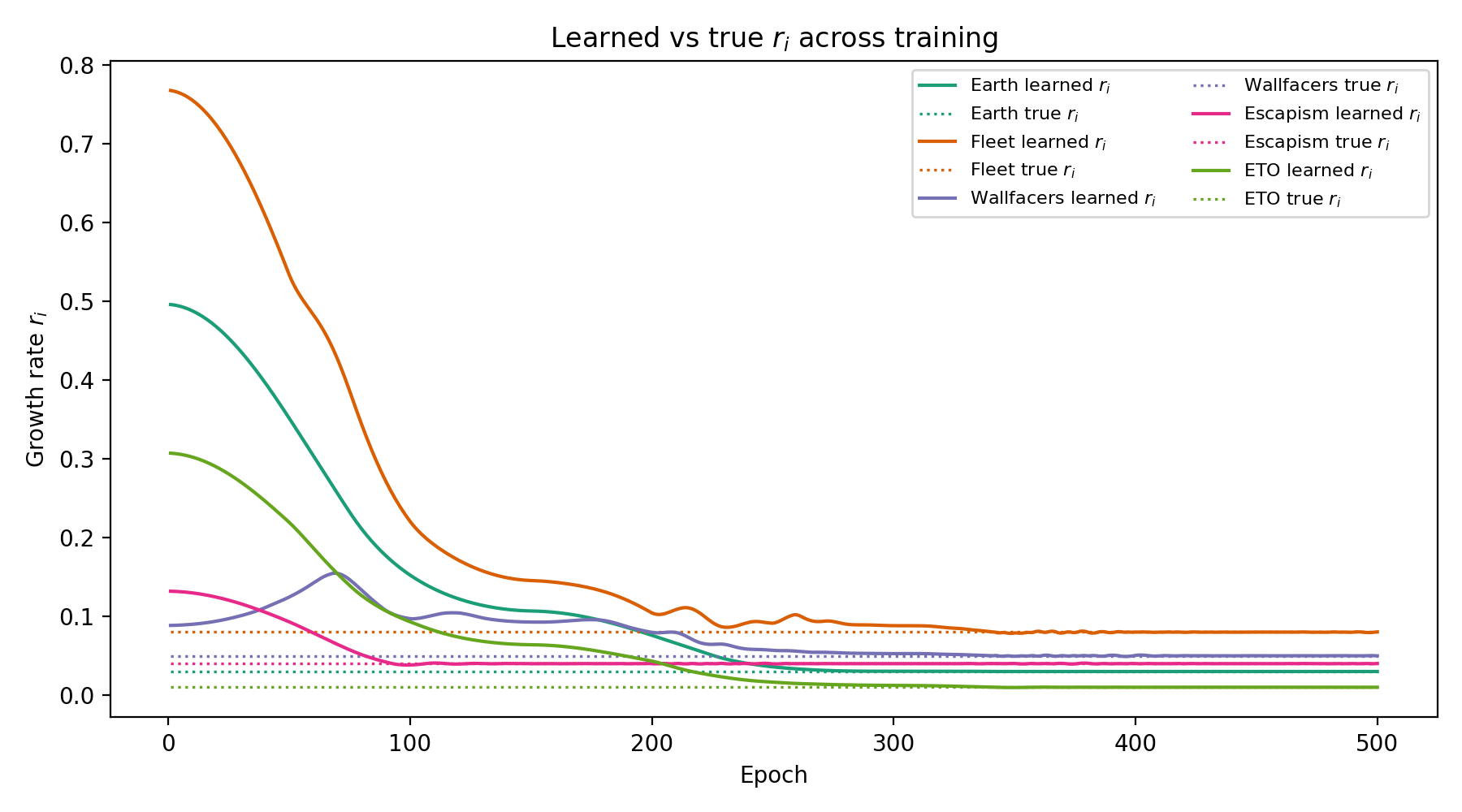}
    \caption{Three-body GLV with explicit self/interaction terms. Top: shared-parameter rollout trained only on $t\in[0,50]$ (green) yet evaluated through $t=250$, including highlighted exogenous-shock regimes and the Sophon lock removal marker. Bottom: learned growth-rate parameters $r_i$ converging from random initialization (solid) toward the ground-truth rates (dotted) during optimization.}
    \label{fig:three-body-explicit}
\end{figure*}

\changed{The experiment in Fig.~\ref{fig:three-body-explicit} uses the following hyperparameters: \(\text{epochs}=500\), base learning rate \(5\times 10^{-4}\) (AdamW with cyclical max learning rate $=2\times 10^{-2}$, cyclical step size $=75$), RK4 rollout horizon \(t_{\max}=250\) with training truncated at \(t_{\text{end}}=50\), curriculum initialized at \(T_{\text{base}}=5\) and incremented by \(4\) every \(50\) epochs, gradient clipping at \(2.0\), and stagnation threshold \(50\) epochs before forcing a curriculum increase.}

\changed{The exogenous forcing $\tau_i(t)$ follows the canonical eras Hysteria $\to$ Great Ravine $\to$ Humanist Recovery $\to$ Post-deterrence Stability.  Each faction receives a piecewise schedule aligned with its narrative arc (Earth governance and the Fleet swing from negative to positive support as the crisis unfolds, Wallfacers hold a milder trajectory, the ETO gains strength post-deterrence). Because $\tau_i(t)$ sits inside the parentheses of Eq.~\eqref{eq:trisolaris-glv}, any regime support or repression shifts the instantaneous growth rate immediately. For Earth governance and the Solar Fleet we use}
\begin{equation}
S_i(t)=
\begin{cases}
S_{i,\text{block}}, & t < t_{\text{deterrence}}\\[4pt]
1.0, & t \ge t_{\text{deterrence}},
\end{cases}
\qquad t_{\text{deterrence}}=180,
\label{eq:SI-trisolaris-s}
\end{equation}
\changed{Fundamental physics remains throttled until deterrence ends: each faction inherits a constant Sophon block value $S_{i,\text{block}}$ (hard-science actors such as the Fleet get lower values than strategy-first factions) and instantly recovers to $S_i=1$ when $t$ reaches $t_{\text{deterrence}}=180$.}
\changed{with $S_{i,\text{block}}\in[0.2,0.7]$ chosen per faction (hard-science actors such as the Fleet receive the strongest restriction, while strategy-first factions like the Wallfacers retain partial scientific capacity).}
\begin{equation}
\begin{aligned}
\tau_{\text{Earth}}(t) &=
\begin{cases}
-0.05, & 0 \le t < 20 \\
-0.15, & 20 \le t < 70 \\
+0.05, & 70 \le t < 180 \\
+0.02, & t \ge 180,
\end{cases}\\[6pt]
\tau_{\text{Fleet}}(t) &=
\begin{cases}
+0.10, & 0 \le t < 20 \\
-0.05, & 20 \le t < 70 \\
+0.02, & 70 \le t < 180 \\
+0.15, & t \ge 180,
\end{cases}
\end{aligned}
\label{eq:SI-trisolaris-tau}
\end{equation}
\changed{Wallfacers, Escapists, and the ETO receive analogous piecewise schedules with signs chosen to match the novels (Escapists benefit during the Ravine, the ETO peaks post-deterrence).  Because $\tau_i(t)$ sits inside the parentheses of Eq.~\eqref{eq:trisolaris-glv}, regime support or repression is translated immediately into the faction's instantaneous growth rate.}

\changed{The interaction matrix $A$ encodes strategic relationships between factions. The diagonal elements $A_{ii}=-0.001$ represent self-limiting effects (negative self-interaction). Off-diagonal elements capture asymmetric interactions: $A_{\text{Fleet},\text{Earth}}=+0.004$ (the Fleet benefits from Earth's economy), $A_{\text{Earth},\text{Fleet}}=-0.005$ (Earth is drained by Fleet investment), $A_{\text{Fleet},\text{ETO}}=-0.01$ and $A_{\text{Wallfacers},\text{ETO}}=-0.015$ (both the Fleet and Wallfacers are actively opposed by the ETO). The matrix is sparse, with most entries zero, reflecting that not all factions directly interact in the narrative.}

\section*{Appendix B: Contagion dynamics -- case study}
\label{sec:appendix-contagion-dynamics}

\begin{figure*}[!t]
    \centering
    \includegraphics[width=0.98\linewidth]{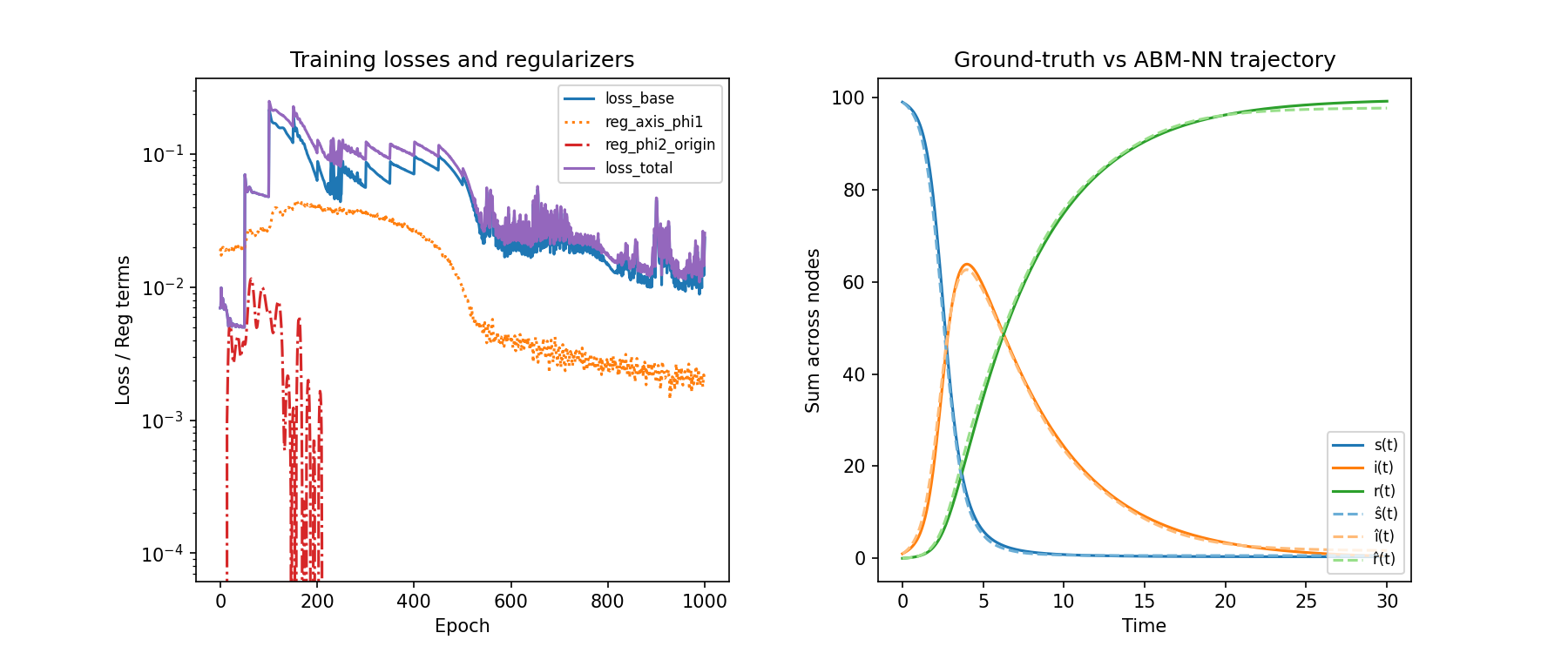}
    % Generated by: python epidemic_demo_macro.py --epochs 1000 --exp_name macro_sir (kept as a note, file name omitted from caption)
    \caption{Training results for the ABM-informed neural network on SIR dynamics with default parameters (\(n=100\), \(p=0.05\), \(\beta=0.4\), \(\gamma=0.2\), \(\mathrm{d}t=0.1\)). The model fits aggregate trajectories while respecting compartmental structure.}
    \label{fig:macro-sir-training}
\end{figure*}

\changed{The run shown in Fig.~\ref{fig:macro-sir-training} uses the following hyperparameters:
Graph \& Simulation Parameters: The simulation uses an Erdős--Rényi graph with $n = 100$ nodes and edge probability $p = 0.05$. The SIR model parameters are infection rate $\beta = 0.4$ and recovery rate $\gamma = 0.2$. The simulation runs for $t_{\max} = 30.0$ time units with integration step size $\Delta t = 0.1$.
Training Parameters: Training runs for $1000$ epochs using the Adam optimizer with learning rate $\eta = 10^{-4}$. Gradient clipping is applied with maximum norm $10.0$. Learning rate decay is disabled ($\gamma_{\text{decay}} = 1.0$).
Network Architecture: Both $\phi_1$ and $\phi_2$ networks use $32$ hidden units per layer and $3$ hidden layers.
Curriculum learning: The training horizon starts at $t_{\text{base}} = 1.0$ and increases by $\Delta t_{\text{horizon}} = 1.0$ every $50$ epochs, up to a maximum training horizon of $t_{\text{train},\max} = 30.0$.
Regularization: Two regularizers are applied: $\lambda_{\phi_1,\text{axis}} = 1.0$ weights the $|\phi_1(S,0)|$ axis regularizer, and $\lambda_{\phi_2,\text{zero}} = 1.0$ weights the $|\phi_2(0)|$ zero regularizer. Regularizers are introduced after a warmup period of $100$ epochs.}

\paragraph*{SIR case study -- noise robustness}
\changed{We assess how the macro-level learner behaves when the aggregate SIR curves are corrupted by observational noise. Using the noise-robustness variant of the learning algorithm, we replay the SIR-on-graph experiment with the same Erdős–Rényi prior ($n=100$, $p=0.05$), base parameters ($\beta=0.4$, $\gamma=0.2$), curriculum schedule (1.0 horizon increment every 50 epochs starting at $t=1$), and the triangular learning-rate policy ($\text{base } lr=10^{-4}$, $\text{max } lr=5\times10^{-4}$, step 100). We vary only the observational noise injected into $s(t),i(t),r(t)$ for $t\geq 3$: Gaussian noise with $\sigma\in\{1.0,2.0,3.0\}$ is added per time step, while the differentiable learner sees only these noisy trajectories. 
Each run trains for 1000–1500 epochs (longer for higher noise).}

\begin{figure*}[!t]
    \centering
    \includegraphics[width=0.8\linewidth]{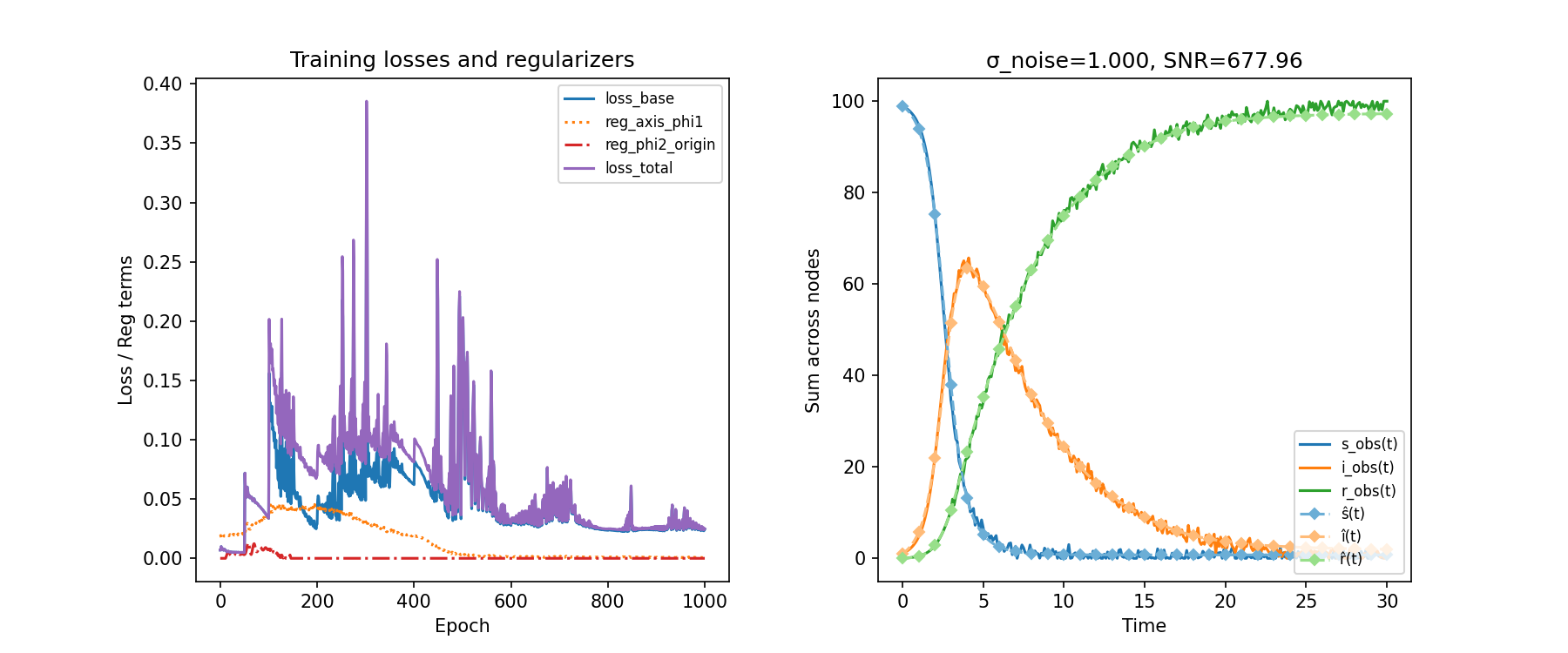}\\[6pt]
    \includegraphics[width=0.8\linewidth]{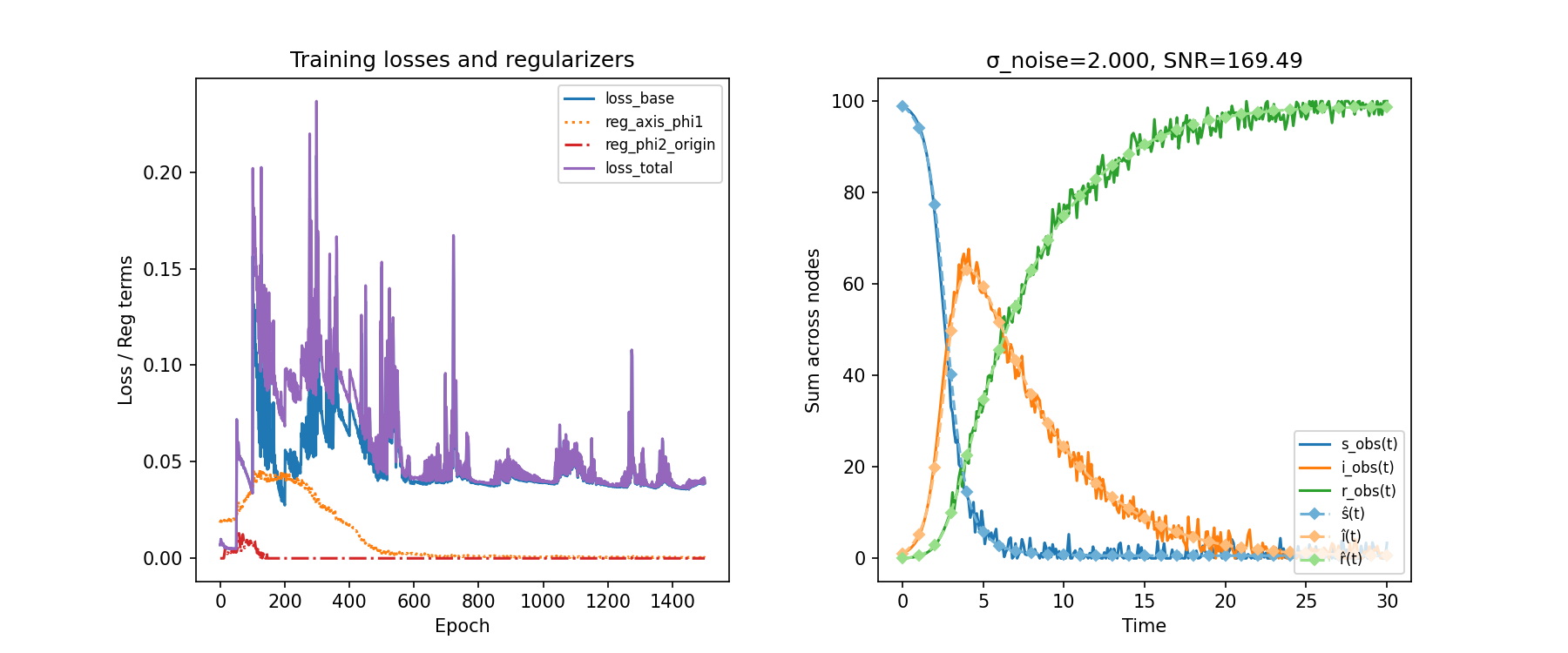}\\[6pt]
    \includegraphics[width=0.8\linewidth]{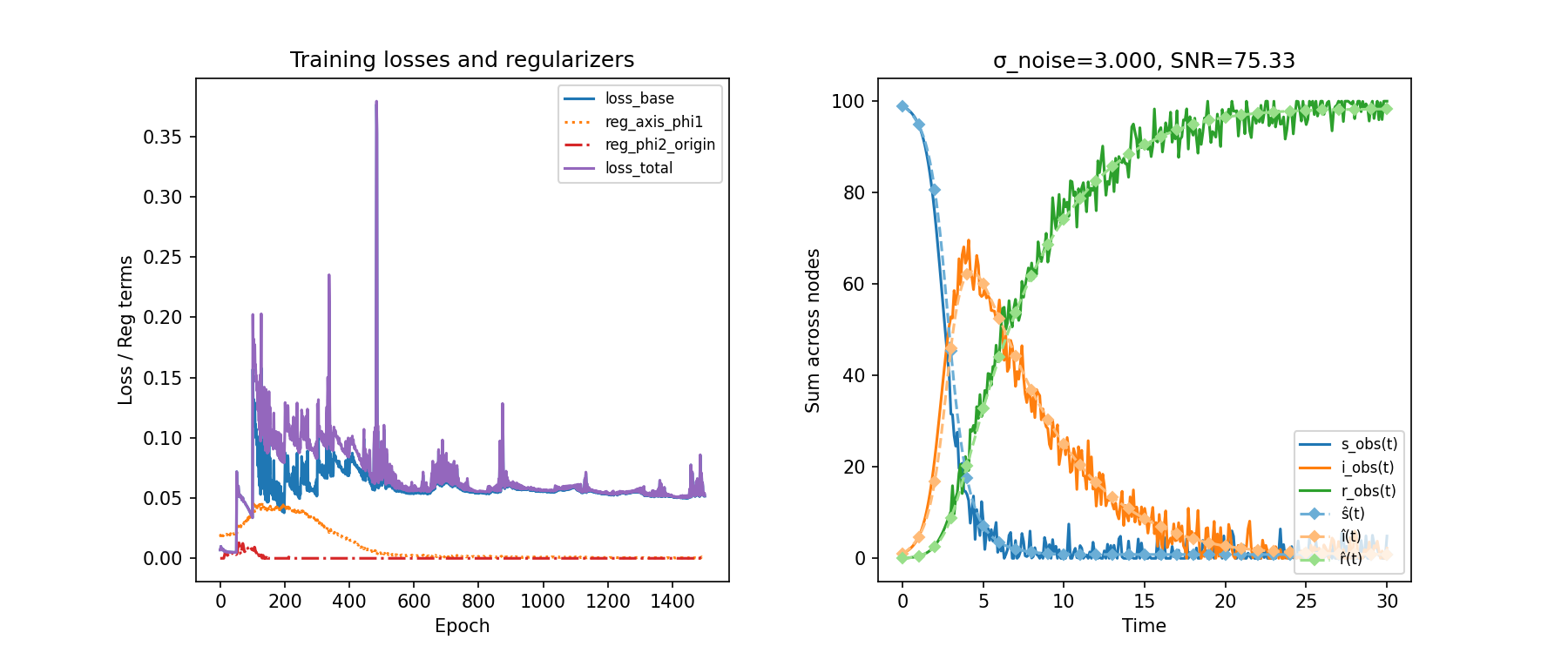}
    \caption{Noise-robustness study for the macro SIR learner. Columns correspond to observational noise $\sigma=1.0$ (up), $2.0$ (middle), and $3.0$ (down). Each panel overlays the true aggregate curves (solid), noisy observations (dotted), and learned trajectories (dashed with diamonds). Despite the increasing corruption after $t=3$, the ABM-NN continues to track the underlying dynamics and maintains separation between in-sample (denoted by the shaded training region) and out-of-sample forecasts. }
    \label{fig:SI-noise-robustness}
\end{figure*}

\begin{figure*}[!t]
    \centering
    \includegraphics[width=0.85\linewidth]{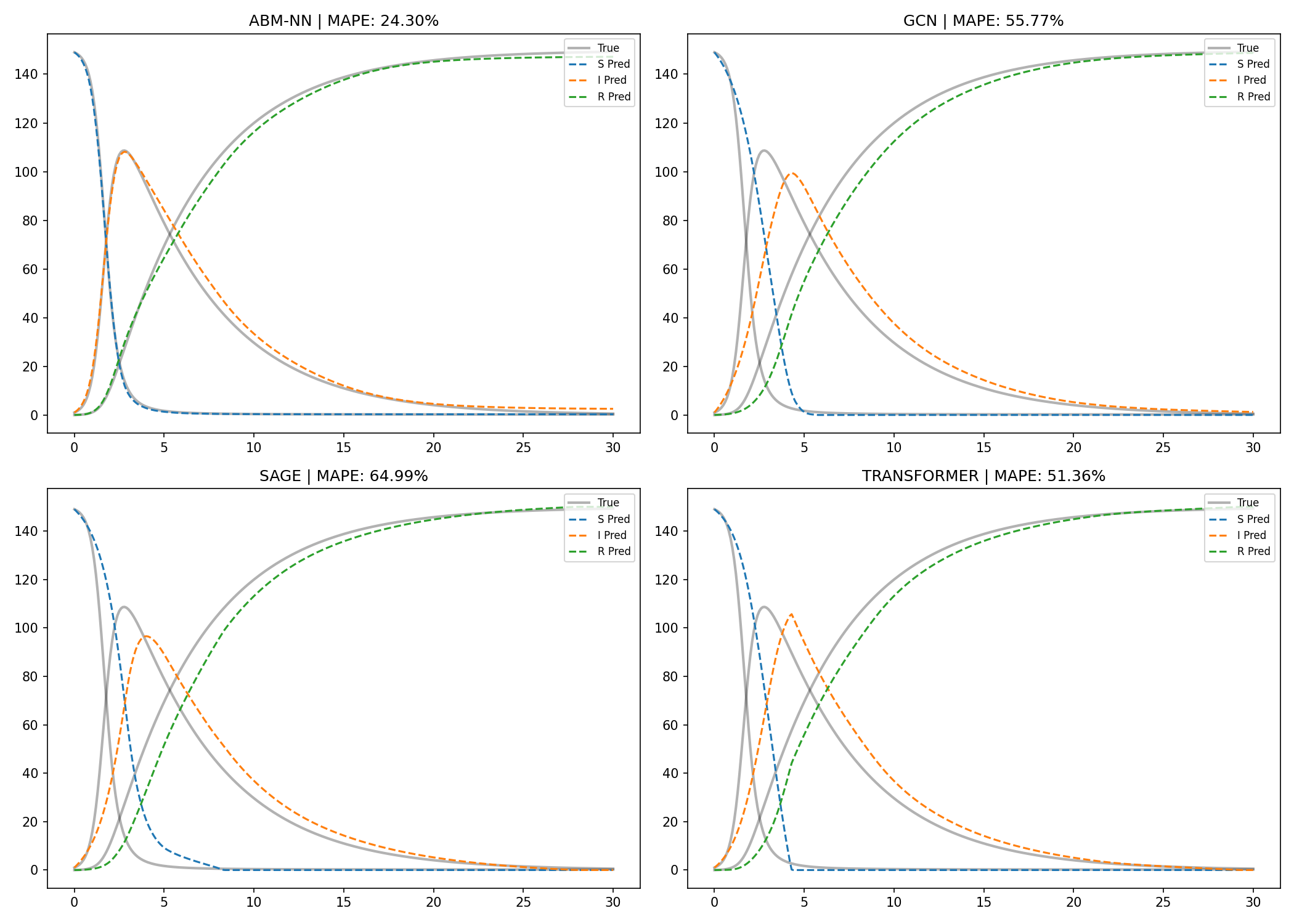}
    \caption{\changed{Comparison of ABM-NN performance against a Graph Convolutional Network (GCN) baseline. The figure displays the predicted trajectories versus ground truth for both models, highlighting the superior performance of ABM-NN in capturing the system dynamics, particularly in out-of-sample scenarios.}}
    \label{fig:baseline-comparison-gcn}
\end{figure*}

\begin{figure*}[!t]
    \centering
    \includegraphics[width=0.85\linewidth]{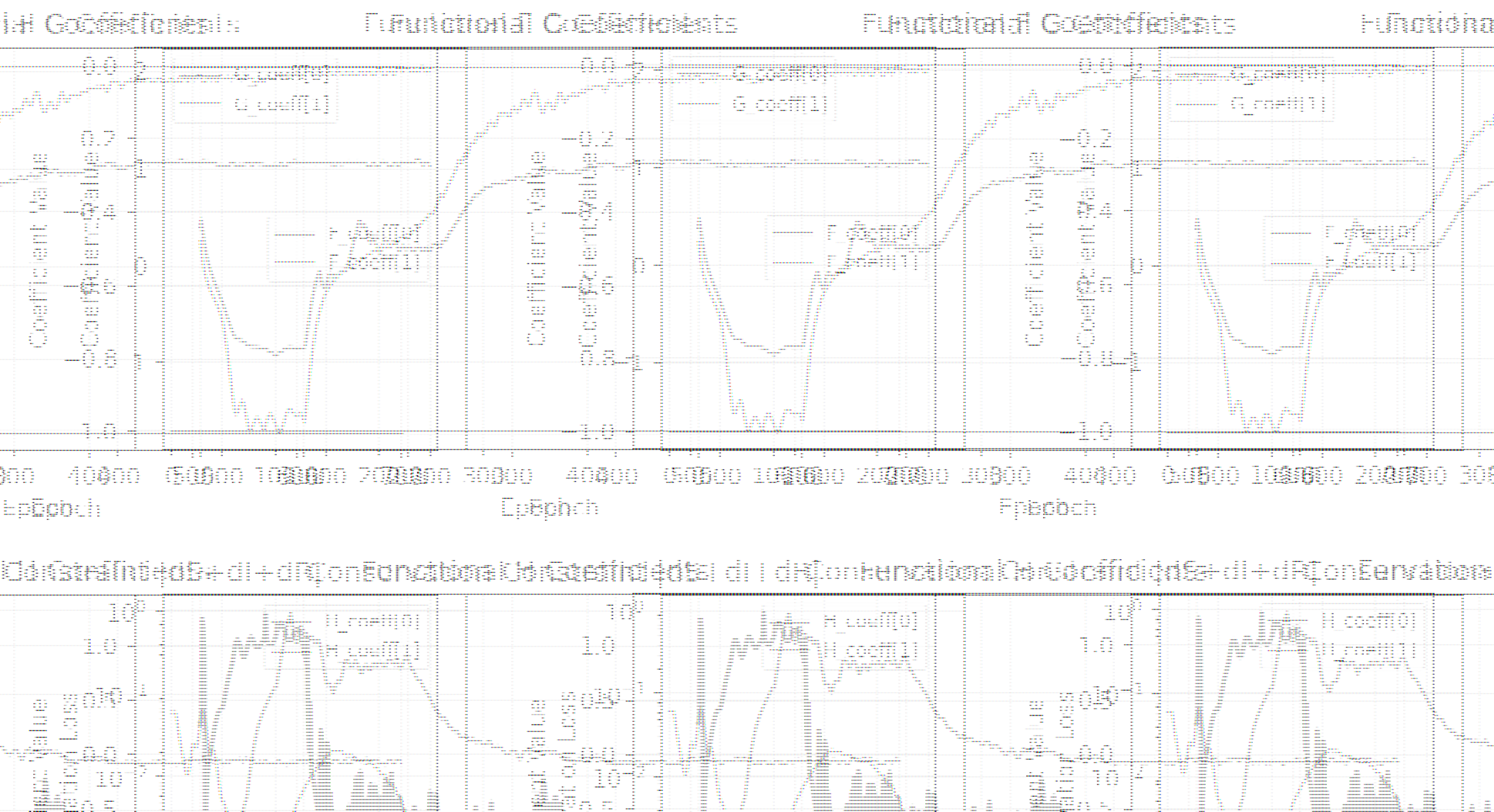}
    \caption{Training results with separate learning rate optimization for different parameter groups. Neural networks ($\phi_1,\phi_2$) use $lr=0.01$ while functional $F$ is held fixed and $G,H$ are trained with $lr=0.1$. Decoupled optimizers and curriculum learning produce smooth gradients, and the learned $G,H$ respect conservation of mass.}
    \label{fig:SI-separate-lr-training}
\end{figure*}

\paragraph*{Functional learning configuration}
\changed{Next, we turn to the functional-learning variant of the SIR study. To keep the interpretation of the functional coefficients anchored, functional $F$ is fixed while $G$ and $H$ are learned; simultaneously, we apply a decoupled learning-rate scheme that assigns separate optimizers to the neural vector field and to each functional coefficient (details and full figure are provided in Fig.~\ref{fig:SI-separate-lr-training}). Curriculum rollout (700 epochs, horizon increment 1.0 every 10 epochs), gradient clipping at 10.0, and LeakyReLU activations stabilize training, while the conservation penalty $\lambda_{\text{conservation}}=10$ keeps $G$ and $H$ mass-preserving. Figure~\ref{fig:SI-separate-lr-training} shows the resulting trajectories together with the optimizer traces, illustrating that the multi-rate scheme eliminates vanishing gradients and keeps $G/H$ aligned with the conservation constraint.}

\clearpage

\section*{Appendix C: Macroeconomic case study}
\label{sec:appendix-macroeconomic-case-study}

\paragraph*{1. State Dynamics}
\changed{The time derivatives for the GDP state $x_i$ and the auxiliary macro state vector $\mathbf{y}_i$ for country $i$ are given by:}
\begin{equation}
\frac{dx_i}{dt} = \underbrace{\phi_x(x_i, \mathbf{y}_i, e_i)}_{\text{Self-Dynamics}} + \underbrace{\psi(x_i, \mathbf{x}, \mathbf{A}, e_i, \mathbf{e})}_{\text{Interaction}}
\end{equation}
\begin{equation}
\frac{d\mathbf{y}_i}{dt} = \phi_y(\mathbf{y}_i, x_i, e_i)
\end{equation}
\changed{where $x_i \in \mathbb{R}$ is the GDP of country $i$, $\mathbf{y}_i \in \mathbb{R}^6$ is the vector of auxiliary macro variables for country $i$, $e_i \in \mathbb{R}^{d_e}$ is the learned static embedding for country $i$, $\mathbf{x}$ is the vector of GDPs for all countries, $\mathbf{A}$ is the learnable adjacency matrix, and $\mathbf{e}$ denotes the set of all country embeddings $\{e_j\}$.}

\paragraph*{2. Neural Parameterizations with FiLM Adapters}
\changed{Both $\phi_x$ and $\phi_y$ use a shared Feed-Forward Network (FFN) backbone adapted via FiLM (Feature-wise Linear Modulation).
For a shared network with $L$ hidden layers, let $\mathbf{h}^{(l)}$ be the activation vector at layer $l$. The forward pass for a single layer $l$ for country $i$ is:}

\subparagraph*{a. Base Linear Transformation:}
\changed{
\begin{equation}
\tilde{\mathbf{h}}^{(l)} = \mathbf{W}^{(l)} \mathbf{h}^{(l-1)} + \mathbf{b}^{(l)}
\end{equation}
}

\subparagraph*{b. FiLM Modulation (Conditioning on $e_i$):}
\changed{The shared adapter networks compute scale ($\gamma$) and shift ($\beta$) parameters from the country embedding $e_i$:
\begin{equation}
\gamma^{(l)}_i = \tanh\left(\mathbf{W}_{\gamma}^{(l)} e_i + \mathbf{b}_{\gamma}^{(l)}\right) + 1
\end{equation}
\begin{equation}
\beta^{(l)}_i = \mathbf{W}_{\beta}^{(l)} e_i + \mathbf{b}_{\beta}^{(l)}
\end{equation}
}

\subparagraph*{c. Modulated Activation:}
\changed{The pre-activation is scaled and shifted element-wise:
\begin{equation}
\mathbf{h}^{(l)} = \text{ReLU}\left( \gamma^{(l)}_i \odot \tilde{\mathbf{h}}^{(l)} + \beta^{(l)}_i \right)
\end{equation}
(where $\odot$ denotes element-wise multiplication).}

\paragraph*{3. Detailed Component Definitions}

\subparagraph*{Self-Dynamics ($\phi_x$):}
\changed{Uses the GLV-inspired structure:}
\begin{equation}
\phi_x(x_i, \mathbf{y}_i, e_i) = x_i \cdot g_\theta(x_i, \mathbf{y}_i, e_i) + d_\theta(x_i, \mathbf{y}_i, e_i)
\end{equation}
\changed{where $g_\theta$ (growth) and $d_\theta$ (drift) are the two scalar outputs of the FiLM-adapted network described above.}

 \subparagraph*{Interaction Term ($\psi$):}
\changed{The interaction term $\psi_i$ for country $i$ is computed as:}
% \begin{equation}
%     \psi_i = \mathcal{C}(k) \cdot s_i \cdot [x_i]_+ \sum_{j} A_{ij} ([x_j]_+ + \epsilon)^\beta,
% \end{equation}
\begin{equation}
    \psi_i = \mathcal{C}(k) \cdot [x_i]_+ \sum_{j} A_{ij} ([x_j]_+ + \epsilon)^\beta,
\end{equation}
where:
\begin{itemize}
    \item $[x]_+ = \max(x, 0)$ ensures non-negative interaction inputs.
    \item $\beta$ is a learnable exponent (initialized to 0.2, constrained to $(0,1)$) that controls the nonlinearity of spillovers.
    \item $A_{ij}$ are the learnable entries of the adjacency matrix $\mathbf{A}$, representing the influence of country $j$ on $i$.
    \item $\epsilon = 10^{-9}$ is a small constant for numerical stability.
    %\item $s_i$ is an optional per-country scaling factor (defaults to 1.0 unless learnable interaction scales are active, in which case $s_i = \tanh(\omega_i) S_{\max}$).
    \item $\mathcal{C}(k)$ is the curriculum interaction factor at epoch $k$, which ramps linearly from 0 to 1 between the warmup and full-coupling epochs to stabilize early training.
\end{itemize}
The optimizer learns the interaction matrix $\mathbf{A}$, the exponent $\beta$.
%and the optional scaling logits $\omega_i$ (if enabled).

\subparagraph*{Auxiliary Dynamics ($\phi_y$):}
\changed{Directly predicts the vector derivative using a separate FiLM-adapted network:}
\begin{equation}
\phi_y(\mathbf{y}_i, x_i, e_i) = \text{NN}_{\phi_y}(\mathbf{y}_i, x_i, e_i)
\end{equation}

\label{sec:macro_hyperparams}
\changed{The experimental results presented in Fig.~\ref{fig:macro-traj-interaction} were obtained using the macro-level training algorithm with the following configuration:}
\begin{itemize}
    \item \textbf{Network Architecture:}
    \begin{itemize}
        \item Both $\phi_x$ and $\phi_y$ utilize $L=2$ hidden layers.
        \item Hidden dimension size $d_\ell = 8$ neurons.
        \item Country embedding dimension $d_e$ implicitly determined by the implementation defaults (typically matched to hidden dimension or small integer like 16).
        \item FiLM adapters applied to all hidden layers.
    \end{itemize}
    
    \item \textbf{Training Schedule:}
    \begin{itemize}
        \item \textbf{Epochs:} 500 total training epochs.
        \item \textbf{Learning Rate:} Base rate $\eta = 5 \times 10^{-4}$ with Cyclical Learning Rate scheduling (max $\eta = 2 \times 10^{-3}$, step size = 75 epochs).
        \item \textbf{Optimization:} Adam optimizer with weight decay $10^{-4}$ and gradient clipping at norm 2.0.
        \item \textbf{Curriculum Learning:} Training horizon starts at $T_{\text{start}}=5$ years, incrementing by 4 years every 10 epochs, with a stagnation check every 50 epochs.
    \end{itemize}

    \item \textbf{Initialization \& Pretraining:}
    \begin{itemize}
        \item \textbf{Pretraining:} 500 epochs of teacher-forcing pretraining (batch size 256, learning rate $10^{-3}$) to initialize $\phi_x$ and $\phi_y$ against finite-difference derivatives before coupling.
        \item \textbf{Interaction Warmup:} Interaction term $\psi$ is introduced gradually, starting after 25 epochs and ramping up over 50 epochs.
    \end{itemize}

    \item \textbf{Regularization \& Constraints:}
    \begin{itemize}
        \item Interaction matrix $\mathbf{A}$ regularization $\lambda_A = 10^{-5}$.
        \item Maximum absolute value of interaction weights $|A_{ij}|$ clamped to 0.6.
        \item Relative error loss $\epsilon = 10^{-2}$ used for robust GDP fitting.
    \end{itemize}
    
    \item \textbf{Data Split:}
    \begin{itemize}
        \item Training period: 1995--2022 (cutoff year).
        \item Evaluation period: 1995--2024.
        \item Top $N=10$ economies by GDP included.
    \end{itemize}
\end{itemize}

\changed{The macro panel used throughout the paper originates from a data utility pipeline, which queries the World Bank API for the ten largest economies (GDP, unemployment, lending interest rate, sovereign debt, working-age population share, CPI inflation, current account balance) and augments the table with the global oil price index from FRED (\texttt{POILAPSPINDEXQ}). 
We then apply a strictly causal constant-velocity Kalman filter to each country/indicator pair using variable-specific process/measurement noise levels; missing values are imputed forward in time without peeking at future observations and the result is stored in the processed dataset. }

\begin{figure*}[!t]
    \centering
    \includegraphics[width=0.7\linewidth]{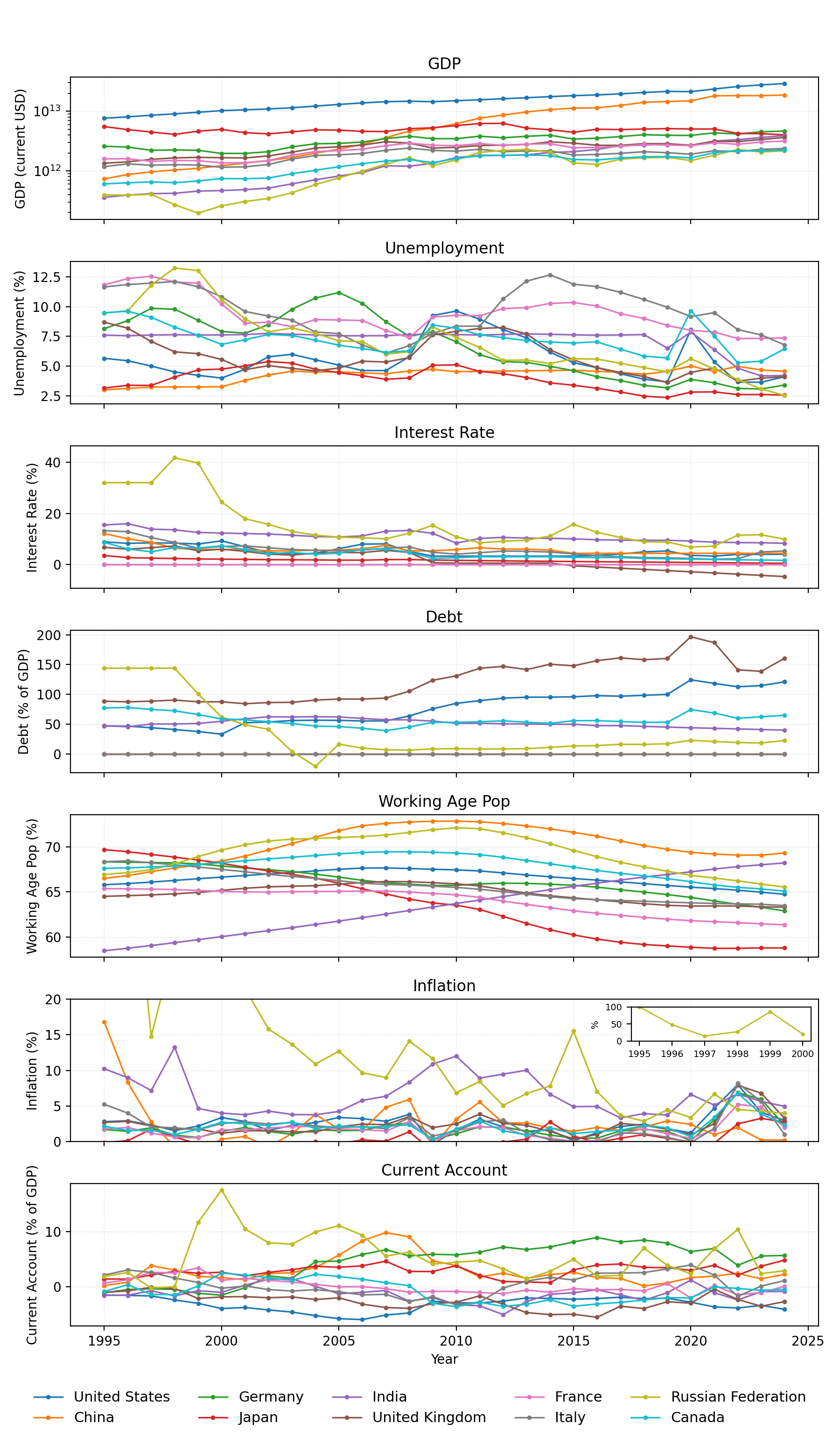}
    \vspace{2mm}
    \caption{Standardized macro covariates (1995--2023) for the ten largest economies. Each series is standardized per country after causal Kalman filtering.}
    \label{fig:SI-macro-covariates}
\end{figure*}

\begin{figure*}[t]
    \centering
    \includegraphics[width=0.75\linewidth]{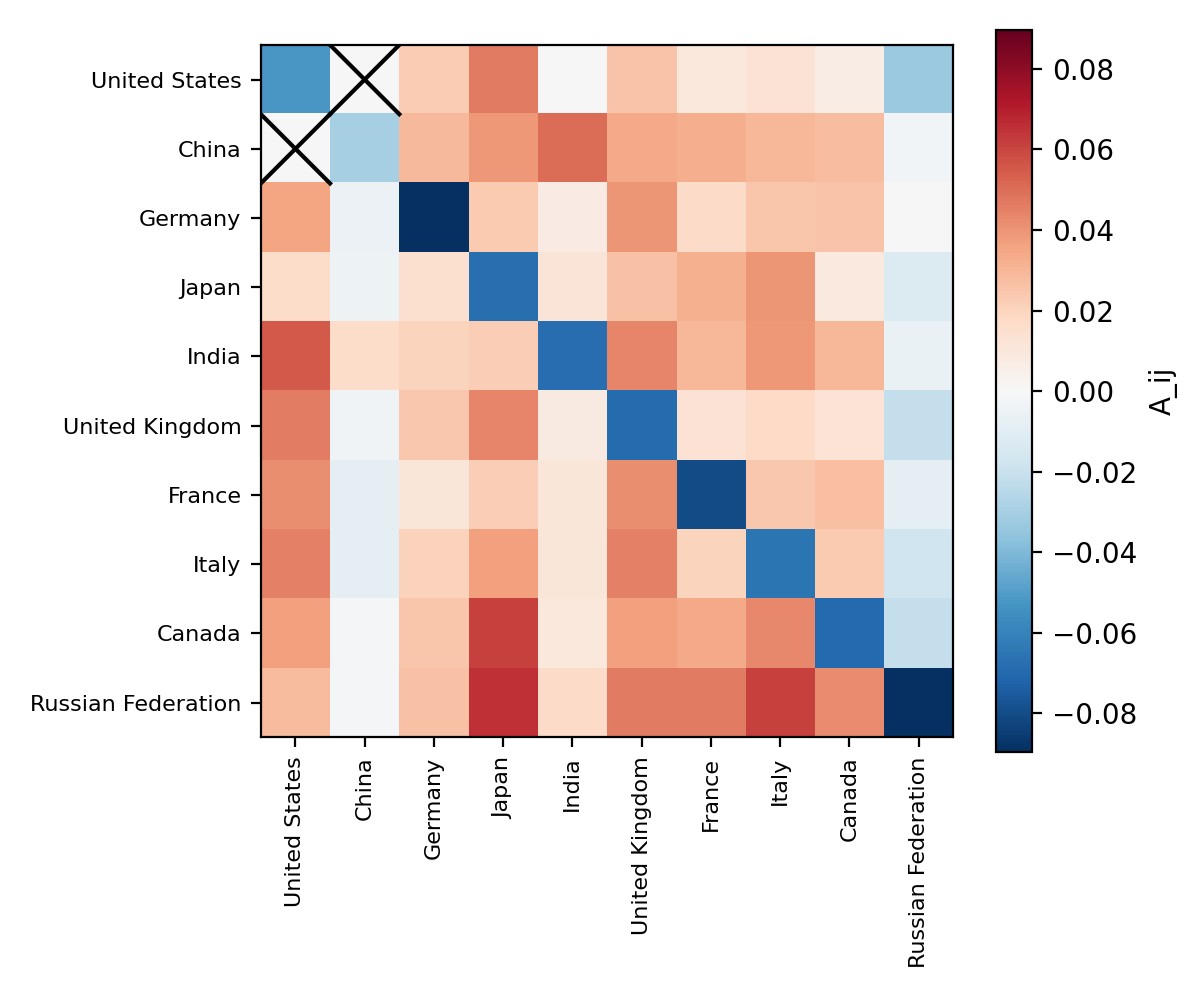}
    \vspace{3mm}
    \includegraphics[width=0.75\linewidth]{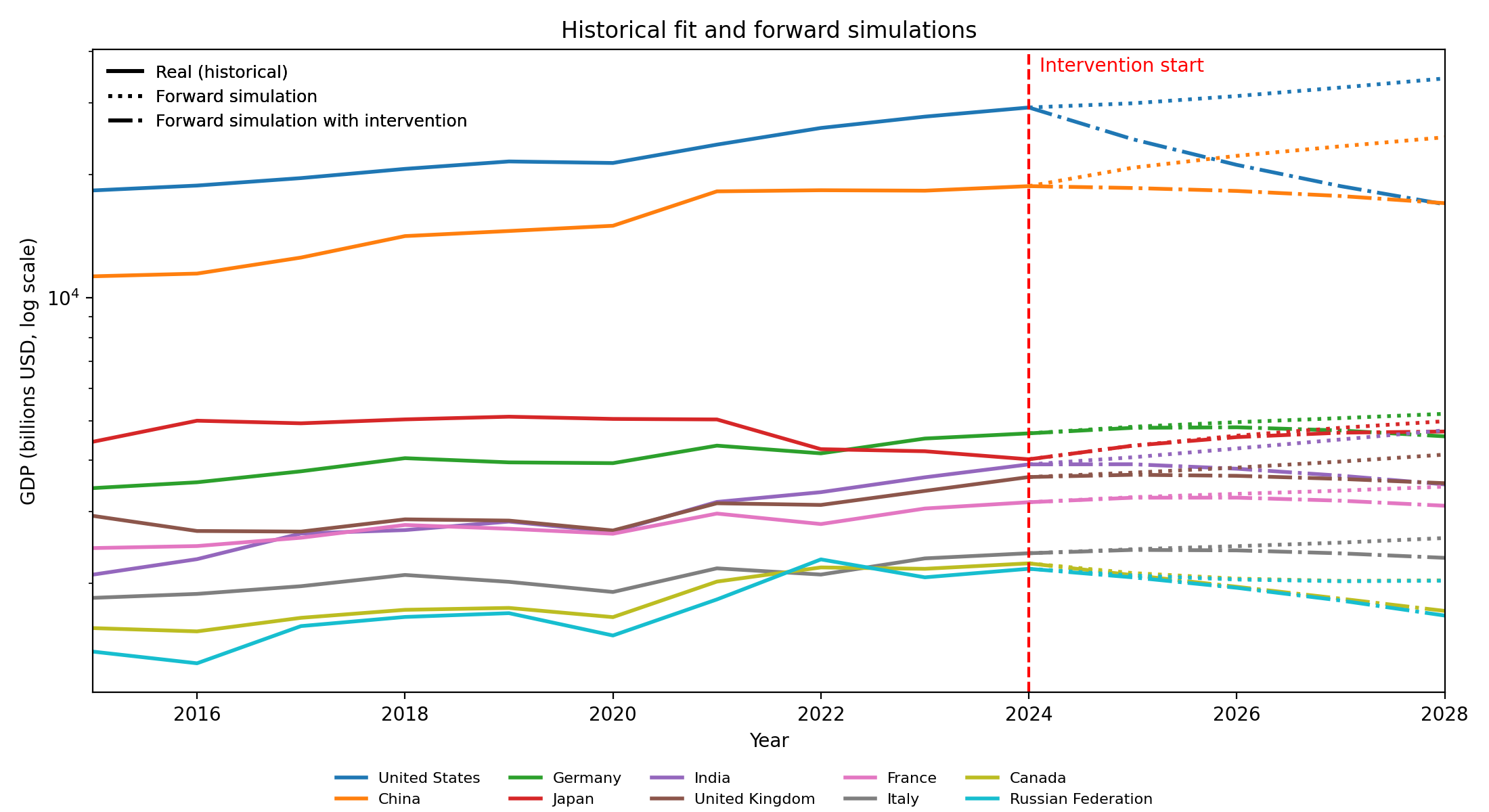}
    \caption{\changed{Counterfactual severing of US--China spillovers. The top panel shows the interaction matrix after zeroing $A_{\text{US},\text{CN}}$ 
    and $A_{\text{CN},\text{US}}$ (highlighted by the black crosses). The bottom panel compares historical GDP (solid), baseline forward simulation (dotted), 
    and the intervention rollout (dash-dot) initialized from the true 2024 state. Removing the bilateral channel (between US and China) dampens post-2024 growth for both economies 
    and lowers the global economy growth for the rest of the world, illustrating how targeted coupling edits propagate through the macroeconomic dynamics.}}
    \label{fig:intervention-simulation}
\end{figure*}

\clearpage

\end{document}